%% file: iclr2027_conference.tex
\documentclass{article} 
\usepackage{iclr2027_conference,times}

\input{math_commands.tex}

\usepackage{booktabs}
\usepackage{hyperref}
\usepackage{url}
\usepackage{amsmath}
\usepackage{amssymb}
\usepackage{mathtools}
\usepackage{amsthm}
\usepackage{subcaption}
\definecolor{mycitecolor}{HTML}{D9A138}
\definecolor{mylinkcolor}{HTML}{D9A138}
\definecolor{myurlcolor}{HTML}{980000}
\definecolor{mydarkgreen}{HTML}{6a994e}
\definecolor{myorange}{HTML}{E7730D}
\definecolor{myblue}{HTML}{4594c1}
\definecolor{sparse1}{HTML}{019E4F}
\definecolor{sparse2}{HTML}{E94127}
\definecolor{sparse3}{HTML}{017DC7}
\definecolor{dense1}{HTML}{3B6C73}
\definecolor{dense2}{HTML}{C1565E}
\definecolor{dense3}{HTML}{4682B4}

\definecolor{mycolor}{RGB}{45, 136, 171} 

\newcommand{\contribution}[1]{%
\begin{tcolorbox}[
    enhanced,
    colback=mycolor!8!white,
    colframe=mycolor,
    leftrule=2mm,
    rightrule=0mm,
    toprule=0mm,
    bottomrule=0mm,
    arc=0mm,
    left=5pt,
    right=5pt,
    top=5pt,
    bottom=5pt,
    breakable,
    leftlower=2mm,
    leftupper=2mm
]
\normalsize 
#1
\end{tcolorbox}
}

\newcommand{\keyquestion}[1]{%
\begin{tcolorbox}[
    enhanced,
    colback=mycolor!8!white,
    colframe=mycolor,
    leftrule=2mm,
    rightrule=0mm,
    toprule=0mm,
    bottomrule=0mm,
    arc=0mm,
    left=5pt,
    right=5pt,
    top=5pt,
    bottom=5pt,
    breakable,
    leftlower=2mm,
    leftupper=2mm,
]
#1
\end{tcolorbox}
}

\usepackage{algorithm}
\usepackage{graphicx}
\usepackage{wrapfig}
\usepackage{algorithmic}
\usepackage{caption}         
\usepackage[most]{tcolorbox} 

\tcbset{
  algobox/.style={
    enhanced,
    boxrule=0.4pt,
    colback=white,
    colframe=black,
    arc=1mm,
    left=6pt, right=6pt, top=4pt, bottom=4pt,
  }
}

\newcommand{\INPUT}{\STATE \textbf{Input:} }
\newcommand{\OUTPUT}{\STATE \textbf{Output:} }
\newcommand{\SET}{\STATE }

\colorlet{lspcolor}{myblue} 
\newcommand{\lsp}[1]{\textcolor{lspcolor}{#1}}
\newcommand{\lspstart}{\begingroup\color{lspcolor}}
\newcommand{\lspend}{\endgroup}

\usepackage[most]{tcolorbox}

\usepackage[normalem]{ulem}
\usepackage{enumitem}
\usepackage{titletoc}

\usepackage{graphicx}
\usepackage{wrapfig}

\usepackage{caption}
\usepackage{multirow}
\usepackage{xcolor}

\usepackage{hyperref}

\title{Latent Space Perturbations Improve Visual Generalization in Reinforcement Learning}

\author{
Daniil Zelezetsky$^{2}$ \quad
Egor Cherepanov$^{1,2}$ \quad
Aleksandr I. Panov$^{1,2}$ \quad
Alexey K. Kovalev$^{1,2}$
\\[0.8em]
$^{1}$AXXX, Moscow, Russia \\
$^{2}$MIRAI, Moscow, Russia
\\[0.5em]
Correspondence: Daniil Zelezetsky \\
\texttt{zelezetsky.d@miriai.org}
}

\iclrfinalcopy 
\begin{document}

\maketitle

\begin{abstract}
Visual generalization in reinforcement learning (RL) means keeping the same task performance when the visuals change, for example the background, lighting, or object appearance, while the dynamics and rewards stay the same. Humans handle such changes well, but RL agents often lose a lot of performance under these visual shifts. Inspired by efficient coding theory from neuroscience, we study how visual shifts change the encoder latent space. We find that encoders trained in fixed visual settings learn a mostly low-dimensional latent space: out-of-distribution (OOD) visuals mainly cause mean and variance shifts along the top principal components of the training latent space, instead of creating new directions. We validate this on our diagnostic environment (LSP-Env) with controlled tests for background changes, distractors, and agent appearance changes, and we also validate this on Procgen. Based on these results, we propose \textbf{Latent Space Perturbation (LSP)}, a PCA-guided method that adds variance-scaled Gaussian noise along dominant latent directions during training. Across LSP-Env, Procgen, and RL-ViGen, LSP improves OOD performance without hurting in-distribution performance, and it does so without changing the encoder or using pixel-level augmentation.
\end{abstract}

\section{Introduction}

Reinforcement learning (RL) agents often generalize poorly to visual conditions unseen during training, even when task dynamics and rewards remain unchanged~\citep{packer2018assessing,kirk2023survey,raileanu2020automatic,li2021domain,wang2025dyn,klepach2025object}. Such failures suggest that learned representations capture task-irrelevant visual factors, causing policies to degrade under changes in background, lighting, or appearance.

Existing approaches mainly rely on input-space augmentation, architectural regularization, or invariant representation learning, but provide limited insight into how OOD visual shifts alter the latent representations mediating perception and control. Understanding these changes in latent statistics and geometry is therefore central to diagnosing visual generalization failures and motivates two questions:

\keyquestion{
    1. How do OOD visual changes deform the geometry and statistics of the encoder latent space?
    
    2. How can training be modified to prevent policy degradation under such distributional shifts?
}

Inspired by the broader intuition of efficient coding~\citep{attneave1954some,barlow1961possible}, we examine whether visual variability in RL representations is concentrated along a small number of latent directions. Empirically, we find that convolutional encoders trained under fixed visual distributions exhibit a low-rank structure, with the leading principal components capturing most of the train-time latent variability.
Under controlled protocols, OOD visual perturbations manifest primarily as mean and variance shifts along existing train-time directions rather than introducing new high-variance components.
We further verify that this geometry persists for task-trained CNNs under naturalistic video-background shifts and for a frozen DINOv2 ViT under mild-to-moderate shifts.
Although the latent marginals are not strictly Gaussian, most of the train-to-evaluation discrepancy is still explained by changes in location and scale, with higher-order effects remaining secondary.

\begin{wrapfigure}[38]{r}{0.5\textwidth}
    \centering
    \vspace{-1.0em}
    \includegraphics[width=0.48\textwidth]{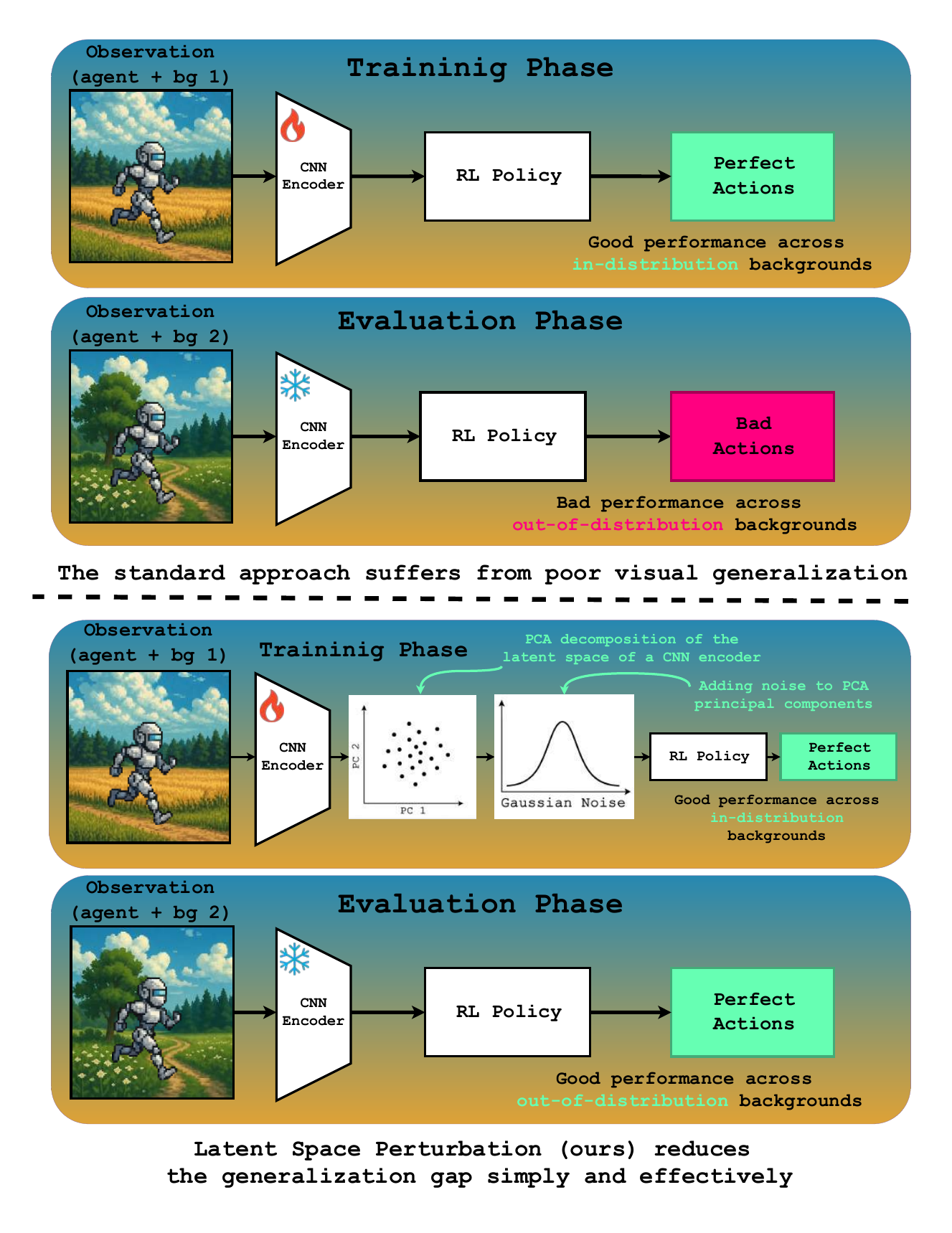}
    \caption{
        \textbf{Latent Space Perturbation (LSP) for visual OOD generalization in RL.}
        \textbf{Top:} Standard vision-based RL pipeline. A CNN encoder and policy trained under a fixed visual distribution perform well in-distribution but fail under evaluation-time visual shifts, as latent representations move outside the region seen during training.
        \textbf{Bottom:} Latent Space Perturbation (LSP). During training, Gaussian noise is injected along PCA-identified latent principal components, training the policy to tolerate dominant visual variation and yielding robust performance under both in-distribution and OOD conditions.}
    \label{fig:visual-abstract}

    \centering
    \includegraphics[width=0.48\textwidth]{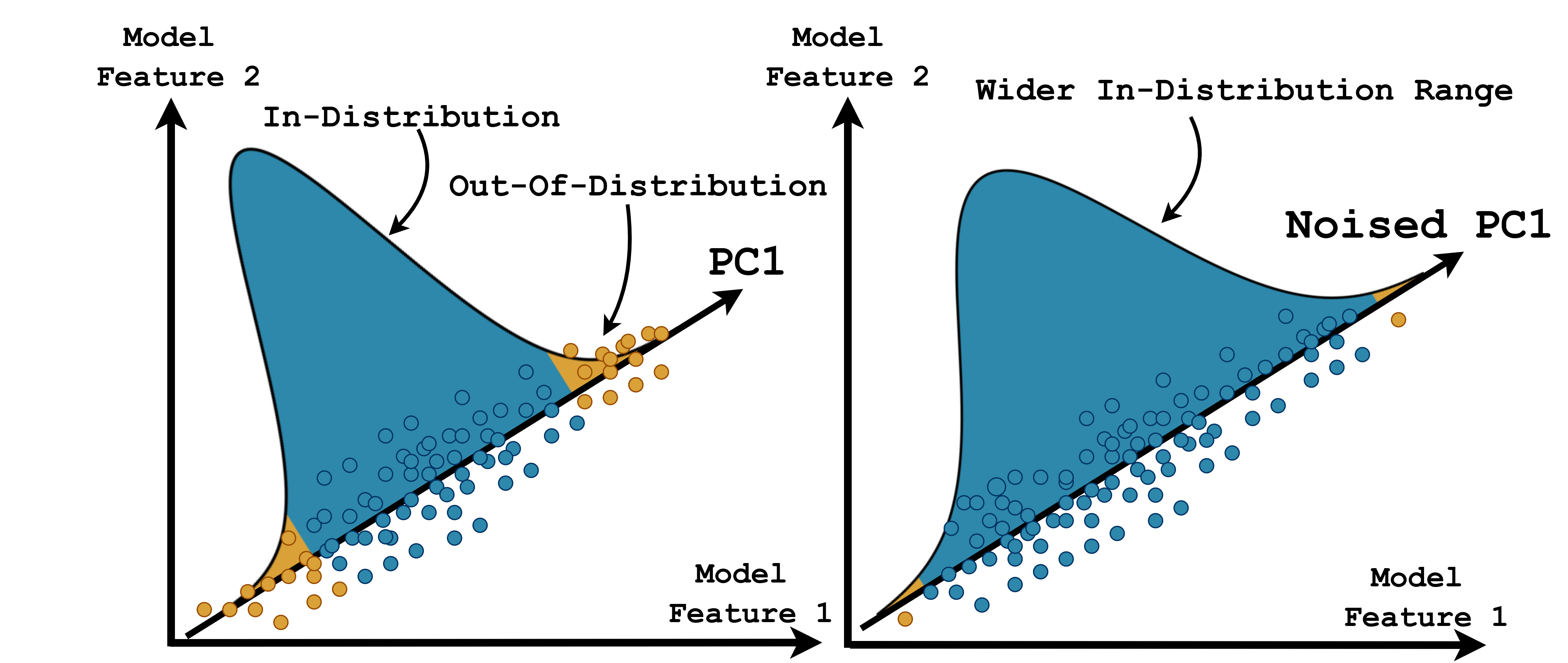}
    \caption{
        \textbf{Geometric intuition for Latent Space Perturbation.}
        \textbf{Left:} OOD observations shift latents along dominant principal components,
        moving outside the in-distribution region and causing poor generalization.
        \textbf{Right:} LSP injects noise along these directions during training, expanding
        in-distribution support and improving robustness.
    }
    \label{fig:scheme}
\end{wrapfigure}

This suggests a direct mitigation. If OOD behavior arises from structured shifts along dominant latent directions, exposing the policy to such perturbations during training should improve inference-time robustness.
We formalize this in \textbf{Latent Space Perturbation (LSP)}, a PCA-guided method that injects variance-scaled Gaussian noise into encoder outputs along leading train-time principal components.
By scaling noise by latent variance, LSP matches the geometry of OOD shifts using latent statistics alone without modifying the encoder or applying pixel-level augmentation.
Importantly, LSP does not require task-relevant and nuisance factors to be disentangled into separate principal components.

To evaluate visual robustness systematically, we build a diagnostic task suite (LSP-Env) with configurable protocols that isolate distinct sources of visual shift: background variation, non-agent distractors, and agent appearance.
Across LSP-Env, Procgen~\citep{procgen}, and RL-ViGen~\citep{rl-vigen}, LSP improves OOD robustness without sacrificing in-distribution performance.
On Procgen, it outperforms the pixel-augmentation baseline RAD on four of five tasks and matches or exceeds IBAC-SNI on four of five tasks at the final checkpoint. RL-ViGen shows that the gains extend to continuous-control robotic manipulation.
The goal is not to learn an explicitly invariant encoder, but to regularize the policy against latent shifts induced by visual changes.

\vspace{-0.5em}
\contribution{
\textbf{We summarize our main contributions as follows:}

\begin{enumerate}[leftmargin=*, itemsep=0em]
    \item We empirically characterize how visual distribution shift affects encoder latents, showing that OOD effects predominantly manifest as structured drift within the leading train-time latent subspace, and verify this geometry across controlled and naturalistic shifts and beyond CNN encoders.

    \item We propose \textbf{Latent Space Perturbation (LSP)}, which injects PCA-aligned noise during training to improve visual robustness.
\end{enumerate}
}

\section{Related Work}
\vspace{-1.0em}

Prior work on visual generalization in RL falls mostly into four directions: (i) observation-space
augmentation, (ii) masking and object-centric generalization, (iii) representation regularization and auxiliary objectives, and (iv)
architectural and optimization-level modifications.

\vspace{-1.0em}
\paragraph{Observation-Space Augmentation.}
RAD~\citep{rad} shows that simple image transformations improve robustness, while automated
methods learn task-specific augmentation policies~\citep{raileanu2020automatic}.
Trajectory-level methods synthesize novel visual inputs via style transfer (Thinker~\citep{thinker})
or environment interpolation (MixReg~\citep{mixreg}).
SVEA~\citep{hansen2021stabilizing} studies instability caused by strong augmentations and improves
stability of Q-learning under augmented observations.
These approaches assume training-time transformations span evaluation-time variation and can
fail under qualitatively different shifts. In contrast, our approach operates in latent space,
perturbing statistically dominant directions rather than applying pixel-level transformations.

\vspace{-1.0em}
\paragraph{Masking and Object-Centric Generalization.}
Several works improve robustness by filtering task-irrelevant visual information before policy learning.
VAI~\citep{wang2021unsupervised} learns distraction-invariant foreground masks through unsupervised visual attention,
while ViSaRL~\citep{liang2024visarl} incorporates human saliency priors into visual representations.
SAM-G~\citep{wang2023generalizable} leverages Segment Anything Model (SAM) for task-relevant object segmentation,
and DeGuV~\citep{pham2025deguv} uses depth masking to suppress irrelevant visual variation.
These methods employ observation filtering, object-centric, and disentangled representations~\citep{kirilenko,kirilenko2023quantized,korchemnyi2026symbolic,vorobyov2026compas},
whereas our work studies how OOD shifts deform latent geometry itself and regularizes the policy directly in latent space.

\vspace{-1.0em}
\paragraph{Representation Regularization and Auxiliary Objectives.}
IBAC-SNI~\citep{igl2019generalization} combines an information bottleneck with selective noise
injection, while CTRL~\citep{ctrl}, ATC~\citep{atc}, and ITER~\citep{igl2} use self-supervised
or contrastive objectives to decouple representation learning from policy optimization.
CLOP~\citep{clop} and architectural scaling choices such as frame stacking and increased channel
capacity also reduce generalization gaps~\citep{procgen,3d_conv}.
Recent work also explored disentangled latent representations for zero-shot generalization~\citep{batra2024zero}.
None of these methods analyze which latent directions drive OOD failures; our approach
directly characterizes these axes and trains the policy to tolerate perturbations along them.

\vspace{-1.0em}
\paragraph{Architectural and Optimization-Level Decoupling.}
PPG~\citep{ppg} decouples actor and critic optimization, while IDAAC~\citep{idaac} adds
auxiliary losses for visual invariance. IPO~\citep{ipo} and REx~\citep{REx} (building on
IRM~\citep{irm}) penalize spurious correlations via multi-domain objectives. These methods
promote invariance through optimization constraints without analyzing how visual shifts
manifest in latent geometry, which is the central focus of our work.

\vspace{-1.0em}
\section{Background}
\vspace{-1.0em}
This section formalizes visual generalization in RL and introduces the notation and assumptions used throughout the paper. We model visual distribution shift as changes in the observation process that preserve underlying task dynamics and rewards, allowing us to isolate failures arising from perception rather than control. This formalization provides the foundation for our latent-space analysis and the method proposed in subsequent sections.

\begin{figure*}[t]
    \centering
    \includegraphics[width=1\linewidth]{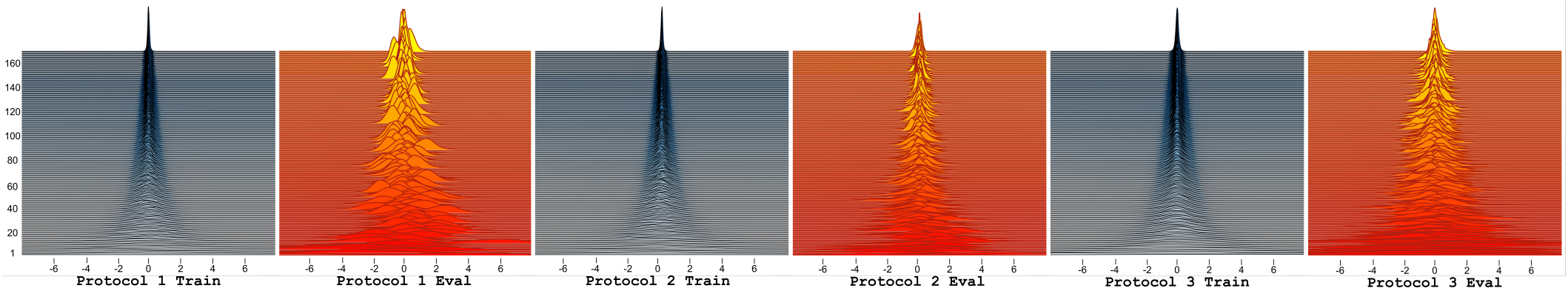}
    \caption{
        \textbf{Plots of latent distributions along the principal components of the train basis.} 
        The blue density functions correspond to the distributions of train latents in the basis formed by the eigenvectors obtained via PCA on the train data itself. 
        The red density functions show the distributions of eval OOD latents projected onto the train basis. 
        Notably, the eval projections remain predominantly single-peaked, while exhibiting shifts in the mean, variance, and higher-order distributional shape.
    }
    \label{fig:joypy_new}
    \vspace{-2.0em}
\end{figure*}

\vspace{-1.0em}
\subsection{Visual Generalization in RL}
\label{subsec:visual_generalization}
\vspace{-1.0em}
We consider a family of partially observable Markov decision processes (POMDPs)
parameterized by an observation function:
$
\mathcal{M}(\Omega)
=
(\mathcal{S}, \mathcal{A}, \mathcal{O}, P, R, \Omega),
$
where $\mathcal{S}$ denotes the \emph{latent environment state space} encoding all
task-relevant variables, $\mathcal{A}$ is the action space, $\mathcal{O}$ the observation space,
$P(s' \mid s,a)$ the state transition kernel, and $R(s,a)$ the reward function.
The observation function
$\Omega : \mathcal{S} \to \mathcal{P}(\mathcal{O})$
maps latent states to probability distributions over visual observations,
where $\mathcal{P}(\mathcal{O})$ denotes the set of probability measures on the observation space $\mathcal{O}$. 
Throughout, we assume all environments in the family share the same $(\mathcal{S}, \mathcal{A}, P, R)$ and differ only in the $\Omega$.
That is, visual variation affects perception but does not alter underlying task dynamics or reward structure.

Let $\Omega_{\mathrm{train}}$ denote the observation function used during training, and let $\Omega_{\mathrm{eval}}$ denote an evaluation-time observation function that differs from $\Omega_{\mathrm{train}}$ only in task-irrelevant visual factors such as texture, background, lighting, or appearance, while preserving the underlying task semantics encoded by $(P,R)$.
We assume the optimal policy $\pi^\star(a \mid s)$ is invariant to observation function $\Omega$.

\begin{wrapfigure}[20]{r}{0.45\textwidth}
\vspace{-1.0em}
    \centering
    \includegraphics[width=0.45\textwidth]{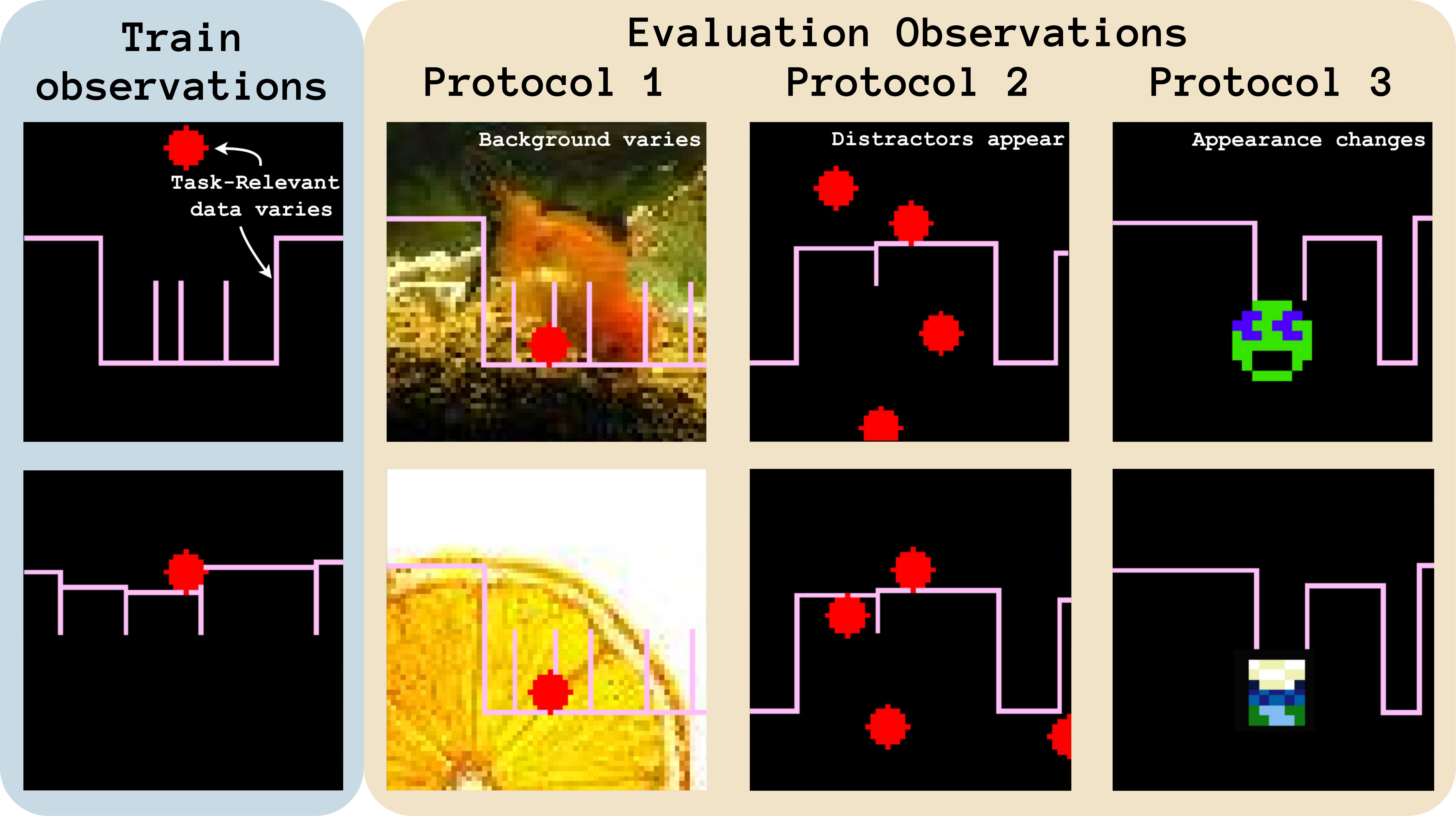}
    \caption{
        \textbf{Train and evaluation observations used for latent space analysis in LSP-Env.}
        \textbf{Train:} Task-relevant factors (agent position and layout) vary while visual appearance (background, distractors, and agent appearance) is fixed.
        \textbf{Eval:} Task-relevant state is fixed and task-irrelevant visual factors varied across three protocols (background, distractors, and appearance), isolating latent variation induced purely by visual changes.
    }
    \label{fig:3plots}
\end{wrapfigure}

For $\pi(a \mid o)$ and $\Omega$, we define the
expected discounted return as
\begin{equation}
J_{\Omega}(\pi)
\;\triangleq\;
\mathbb{E}_{\substack{
s_0 \sim \rho,\\
s_{t+1} \sim P(a \mid s_t,a_t),\\
o_t \sim \Omega(o \mid s_t),\\
a_t \sim \pi(a \mid o_t)
}}
\left[
\sum_{t=0}^{\infty} \gamma^t R(s_t,a_t)
\right]
\end{equation}
where $\rho$ denotes the initial state distribution.
The training objective is $\pi^\star\in\arg\max_{\pi}J_{\Omega_{\mathrm{train}}}(\pi)$.

\vspace{-1.0em}
\paragraph{Definition (Visual Generalization).}
A policy $\pi$ is said to \emph{generalize visually} if its expected return remains unchanged when the observation function is changed in task-irrelevant visual aspects.
Formally, given a training observation function $\Omega_{\mathrm{train}}$ and an evaluation-time observation function $\Omega_{\mathrm{eval}}$ that preserves the same underlying dynamics and rewards $(P,R)$, we define the visual generalization gap as
\begin{equation}
\label{eq:visual_gap}
\Delta_{\mathrm{vis}}(\pi)
\;\triangleq\;
\bigl|
J_{\Omega_{\mathrm{train}}}(\pi)
-
J_{\Omega_{\mathrm{eval}}}(\pi)
\bigr|.
\end{equation}
The policy $\pi$ is said to generalize visually if $\Delta_{\mathrm{vis}}(\pi)
\le \varepsilon$ for a small $\varepsilon > 0$.

\vspace{-1.0em}
\paragraph{Task Formalization.}
The agent implements a policy $\pi(a \mid o)$ that operates on visual observations
through an intermediate latent representation produced by a visual encoder
$f : \mathcal{O} \to \mathbb{R}^D$.
That is, action selection depends on observations only through
$z = f(o)$.
Since the optimal policy in latent state space, $\pi^\star(a \mid s)$, is invariant
across observation functions $\Omega$, the learning problem is perceptual rather
than dynamical.
Failures of visual generalization arise when visually induced observation shifts
cause the latent distribution induced by $z = f(o)$ to move outside the high-density
region encountered during training: $o \sim \Omega_{\mathrm{train}}(o \mid s)
\;\longrightarrow\;
o' \sim \Omega_{\mathrm{eval}}(o \mid s).$

\vspace{-1.0em}
\paragraph{Problem Statement.}
Given a fixed encoder-policy architecture trained under 
$\Omega_{\mathrm{train}}$, our objective is to design a training procedure that reduces the visual generalization gap $\Delta_{\mathrm{vis}}(\pi)$ defined in~\autoref{eq:visual_gap}, without modifying the transition dynamics, reward function, or encoder architecture.
In this work, we study how visually induced OOD shifts from $\Omega_{\mathrm{eval}}$  manifest in the geometry and statistics of the encoder's latent space created by latents $z = f(o')$ and study how visually induced OOD shifts manifest in the geometry and statistics of the encoder's latent space, with the goal of designing training procedures that stabilize the return functional $J_{\Omega}(\pi)$ under such shifts.

\vspace{-1.0em}
\section{Benchmarking Visual Generalization}
\label{sec:environments}
\vspace{-1.0em}

To examine the hypothesis that visually induced distribution shifts exhibit a structured geometry in encoder latent space, we construct a simple diagnostic environment, \textbf{LSP-Env} (\autoref{fig:3plots}).
LSP-Env is a lightweight 2D platformer in which the agent moves right as quickly as possible while avoiding obstacles, with fixed dynamics and objectives and controllable appearance factors.
Observations are $64\times 64$ RGB images and actions are 2D continuous displacements (left/right movement speed and jump force).
The reward encourages forward progress and severely penalizes vertical motion, inducing a consistent ``move-right'' objective.
Within this fixed task, we can isolate task-irrelevant visual variation by independently changing the background image, adding non-controllable agent-like distractors, and altering the controlled agent's appearance (texture).
Details are in~\autoref{app:LSP-Env-details}.

We then evaluate the resulting method on established visual generalization benchmarks to test whether conclusions from the controlled setting transfer to more complex, less controlled environments.
\textbf{Procgen}~\citep{procgen} provides procedurally generated environments with substantial visual diversity, serving as a stress test under uncontrolled appearance variation at scale.
\textbf{RL-ViGen}~\citep{rl-vigen} comprises popular environments adapted for visual generalization evaluation; we use the Robosuite Door and Lift tasks to assess LSP in a challenging continuous-control robotics setting.
These environments also structure the remainder of the work.
In~\autoref{sec:latent-space-analysis}, we use \textbf{LSP-Env} to characterize how appearance-only shifts change the encoder latent distribution and to extract constraints on the kinds of latent perturbations the policy should tolerate.
In~\autoref{sec:lsp_method}, we instantiate these constraints as \textbf{Latent Space Perturbation (LSP)}, a training-time regularizer at the representation-to-control interface.
Finally, \autoref{sec:results} evaluates whether LSP reduces the visual generalization gap on LSP-Env and whether the gains transfer to Procgen and RL-ViGen.

\vspace{-1.0em}
\section{Latent Space Analysis}
\label{sec:latent-space-analysis}
\vspace{-1.0em}
This section analyzes how appearance-only shifts change the encoder latent distribution when dynamics and rewards are fixed, with the aim of motivating the training procedure in~\autoref{sec:lsp_method}.
We measure three effects: shifts in the latent mean, changes in the variance spectrum, and drift of the dominant PCA subspace.
These diagnostics distinguish two regimes: OOD variation that remains within the dominant train-time subspace (suggesting extrapolation by the policy head), versus OOD variation that introduces new dominant directions (suggesting a representation-level limitation).

\vspace{-1.0em}
\subsection{Controlled evaluation protocols}
\label{sec:lsp_protocols}
\vspace{-1.0em}
We use three \textbf{LSP-Env} protocols that vary visual appearance while keeping the task state fixed (see~\autoref{fig:3plots}).
They probe scene-context invariance (background variation), disambiguation under visually similar distractors (agent-like distractors), and stability under changes to the controlled entity's visuals (agent appearance changes).
This separation is aligned with our diagnostics: whether OOD induces low-dimensional translation/rescaling in latent space, or introduces changes that would require updating the representation rather than the policy head.
The agent uses a CNN encoder $f_\theta$ and an MLP policy head (details in~\autoref{app:arch-details}).
Each observation maps to $z=f_\theta(o)\in\mathbb{R}^D$, the interface between perception and control, and we analyze how appearance-only perturbations change the distribution of $z$.

\begin{wrapfigure}[10]{r}{0.5\textwidth}
    \vspace{-0.8em}
    \centering
    \includegraphics[width=0.5\textwidth]{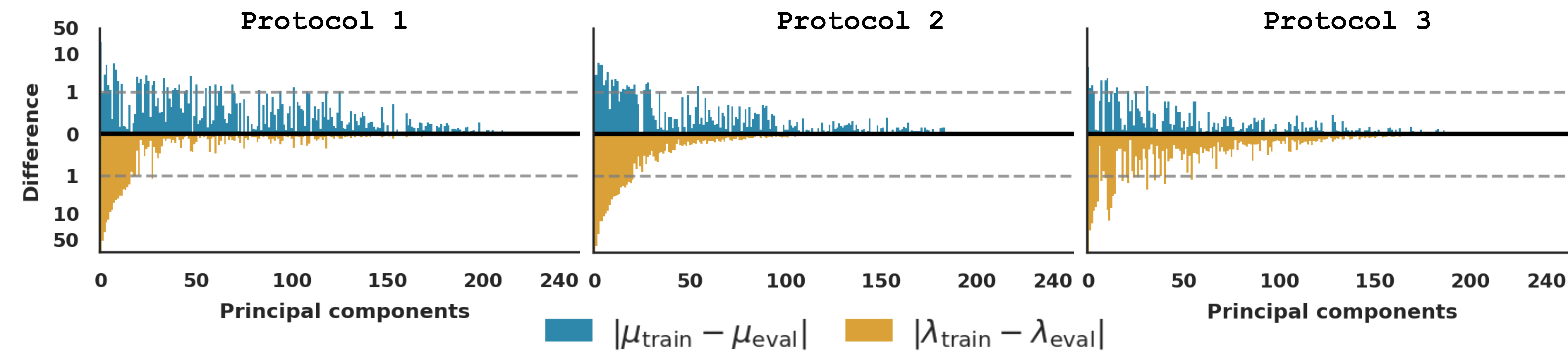}
    \vspace{-1.3em}
    \caption{
        \textbf{Per-component mean and variance shifts (train PCA basis) for three protocols.}
        Top bars: $|\mu_i^{\mathrm{train}} - \mu_i^{\mathrm{eval}}|$; bottom bars: $|\lambda_i^{\mathrm{train}} - \lambda_i^{\mathrm{eval}}|$.
        Linear scale below dashed threshold, log scale above.
        Shifts concentrate in leading PCs; tail components contribute negligibly.
    }
    \label{fig:vis_diff}
    \vspace{-1.2em}
\end{wrapfigure}

\vspace{-1.0em}
\subsection{Data collection with a frozen encoder}
\label{sec:lsp_data_collection}
\vspace{-1.0em}
We analyze a frozen encoder $f_\theta$ trained under $\Omega_{\mathrm{train}}$ and collect latents under controlled train and eval observation processes with encoder parameters fixed.

\vspace{-1em}
\paragraph{Train latents.}
We collect
$
Z_{\text{train}} \in \mathbb{R}^{N \times D}
$
from $N$ observations $o \sim \Omega_{\mathrm{train}}(o \mid s)$ where task-relevant factors vary (agent position, layout, etc.) and appearance is fixed.
In particular, training observations include no background changes, no agent-like distractors, and no changes to the agent appearance ( \autoref{fig:3plots}).

\vspace{-1em}
\paragraph{Eval latents.}
We collect
$
Z_{\text{eval}} \in \mathbb{R}^{N \times D}
$
from observations $o' \sim \Omega_{\mathrm{eval}}(o \mid s)$ where the latent state is held fixed and only task-irrelevant visual factors vary according to three protocols from \autoref{sec:lsp_protocols}.
Thus, differences between $Z_{\text{train}}$ and $Z_{\text{eval}}$ isolate visual OOD effects.

\vspace{-1.0em}
\subsection{Train PCA coordinate system}
\vspace{-1.0em}
\label{sec:lsp_train_basis_choice}

We next define the coordinate system used for all reported train--eval shift metrics.

\vspace{-1em}
\paragraph{Step 1: train PCA basis.}
We apply PCA to $Z_{\text{train}}$ to obtain an orthonormal basis of principal directions and their associated variances:
$
V_t = (v^{t}_{1}, \ldots, v^{t}_{D}), 
\qquad 
\Lambda_t = (\lambda^{t}_{1}, \ldots, \lambda^{t}_{D}),
$
where $v_i^t$ is the $i$-th train principal direction and $\lambda_i^t$ is the variance of train latents along that direction.

\vspace{-1em}
\paragraph{Step 2: project eval onto the train basis.}
Using the same frozen encoder, we collect the eval latents and project them onto the train basis $V_t$ and compute the resulting per-component variances:
$
\Lambda_e = (\lambda^{e}_{1}, \ldots, \lambda^{e}_{D}),
$
where $\lambda_i^e$ is the variance of eval latents along the train direction $v_i^t$.

\begin{figure*}[t!]
    \vspace{-1em}
    \centering
    \includegraphics[width=1\linewidth]{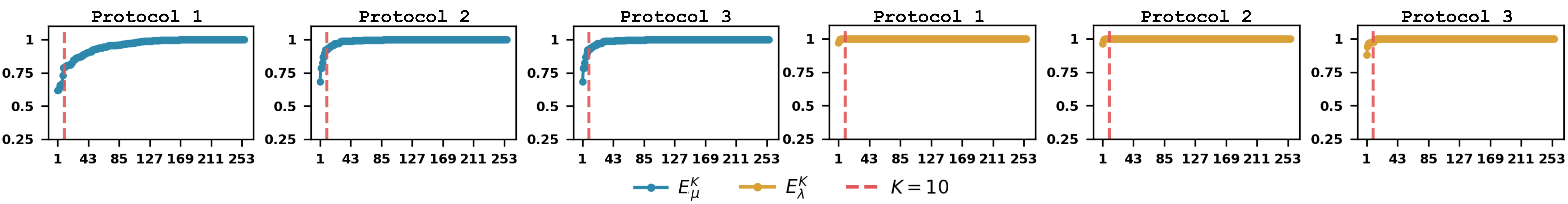}
    \vspace{-1.5em}
    \caption{
        Cumulative fraction of OOD shift energy captured by the top-$K$ train PCs.
        \textbf{Left:} mean-shift $E^K_\mu$; 
        \textbf{Right:} variance-shift $E^K_\lambda$. Both saturate quickly; $K\!=\!10$ already explains nearly all shift.
    }
    \vspace{-1.7em}
    \label{fig:E_mu}
\end{figure*}

Analyzing eval latents in the train PCA basis tests whether OOD appearance changes mainly redistribute mean and variance along directions that dominate training.
If variability stays in the leading train components, the policy head remains closer to its training support; if substantial variance appears orthogonal to the train subspace, eval latents require extrapolation.
Thus, the train PCA basis separates reweighting of existing structure from new dominant directions.
Because this analysis could be misleading if eval induces a rotated dominant subspace, we explicitly verify train--eval subspace alignment in~\autoref{sec:subspace_sim}.
This check ensures that train-coordinate metrics reflect latent geometry rather than a coordinate artifact.

\vspace{-0.5em}
\contribution{
\textbf{This analysis addresses three research questions (RQs):}
\begin{enumerate}[leftmargin=*, itemsep=0.1em]
    \item \textbf{(RQ1) Training latent structure.}
    What statistical and geometric properties characterize the encoder latent space on in-distribution data?

    \item \textbf{(RQ2) Distraction-induced shift.}
    How do these statistics change when OOD visual distractors are introduced?

    \item \textbf{(RQ3) Subspace stability.}
    Does the train PCA subspace remain sufficient to explain eval-time variability, or do OOD observations introduce new latent directions?
\end{enumerate}
\vspace{-0.5em}
}

To answer these RQs, we introduce metrics that quantify train--eval shifts at different levels.

\vspace{-1.0em}
\subsection{Component-level shift metrics in the train basis}
\label{sec:lsp_component_metrics}
\vspace{-0.5em}
Let $Z_{\text{train}}, Z_{\text{eval}} \in \mathbb{R}^{N \times D}$ denote the matrices of train and eval latents collected as described in~\autoref{sec:lsp_data_collection}. All metrics in this subsection are computed in the train PCA basis $V_t$.

\vspace{-0.7em}
\paragraph{Mean shift.}
For each train principal direction $v_i^t$, we compute the coordinate-wise means $\mu_i^t$ and $\mu_i^e$ of the projected train and eval latents. We define the mean-shift vector:
$\Delta \mu = (\mu^e_1 - \mu^t_1, \ldots, \mu^e_D - \mu^t_D),$
and the global mean-shift energy:
$E_\mu = \|\Delta \mu\|^2 = \sum_{i=1}^D (\Delta \mu_i)^2.$

\vspace{-0.7em}
\paragraph{Variance shift.}
Similarly, we define the variance difference vector:
$\Delta \lambda = (\lambda^e_1 - \lambda^t_1, \ldots, \lambda^e_D - \lambda^t_D),$
and the global variance-shift energy:
$E_\lambda = \|\Delta \lambda\|^2 = \sum_{i=1}^D (\Delta \lambda_i)^2.$

\begin{wrapfigure}{r}{0.5\textwidth}
    \vspace{-1.2em}
    \centering
    \includegraphics[width=0.5\textwidth]{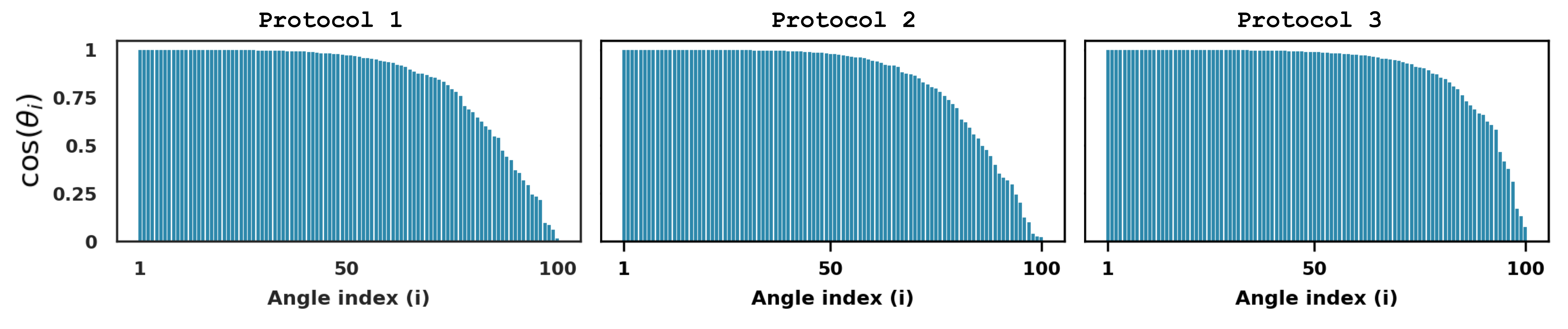}
    \vspace{-0.6em}
    \caption{
        \textbf{Cosines of principal angles between train and eval PCA subspaces.}
        Values near one indicate strong alignment, with deviations only at high-index, low-variance components.
    }
    \label{fig:angles}
    \vspace{-2.0em}
\end{wrapfigure}

These energies provide coarse but informative measures of the overall magnitude of distributional shift induced by OOD visual perturbations. In addition, we compute complementary per-component metrics (mean differences, variance differences, variance fractions, and ratios) in~\autoref{app:metrics} and summarize them in~\autoref{tab:latent_shift_metrics}.

\vspace{-0.5em}
\paragraph{Energy concentration in the dominant train directions.}
To test whether OOD effects concentrate in the top train PCs, we define a set of the top-$K$ train PCs by training variance: $\mathrm{Top\text{-}}K(\lambda^t)$.
We then define the \emph{top-$K$ mean-shift energy ratio} and \emph{top-$K$ variance-shift energy ratio}:
\begin{equation}
\small
    E^K_\mu = \frac{\sum_{i \in \mathrm{Top\text{-}}K(\lambda^t)} (\Delta \mu_i)^2}{E_\mu},
E^K_\lambda = \frac{\sum_{i \in \mathrm{Top\text{-}}K(\lambda^t)} (\Delta \lambda_i)^2}{E_\lambda}.
\end{equation}
These ratios answer whether the dominant train subspace explains most of the shift magnitude.

\vspace{-1.0em}
\subsection{Eval distribution analysis in the train PCA basis}
\label{sec:eval_dist_anal}
\vspace{-1.0em}
We now describe how the eval distribution differs from train when expressed in the train PCA basis.

\vspace{-1.0em}
\paragraph{Empirical shape of 1D coordinate distributions.}
Empirically, the per-component coordinate distributions in the train PCA basis are close to unimodal for most components (see \autoref{fig:joypy_new}).
Accordingly, we summarize the shift using coordinate-wise means and variances, a Wasserstein decomposition confirms that these explain most of the train--eval discrepancy, while the remaining residual reflects higher-order shape changes rather than new modes (see~\autoref{app:higher_order_shift}).
Across all metrics and all three protocols, distractor-induced OOD effects appear primarily as:
(i) systematic mean shifts and (ii) variance redistribution \emph{within the dominant train subspace} (see \autoref{fig:vis_diff}).
Intermediate and tail components can exhibit large ratio-based metrics when $\lambda_i^t$ is near zero, but their contribution to total shift energy is negligible.
Therefore, OOD observations do not substantially change the dominant latent structure learned under $\Omega_{\mathrm{train}}$.

\begin{figure*}[t!]
    \vspace{-0.5em}
    \centering
    \includegraphics[width=1\linewidth]{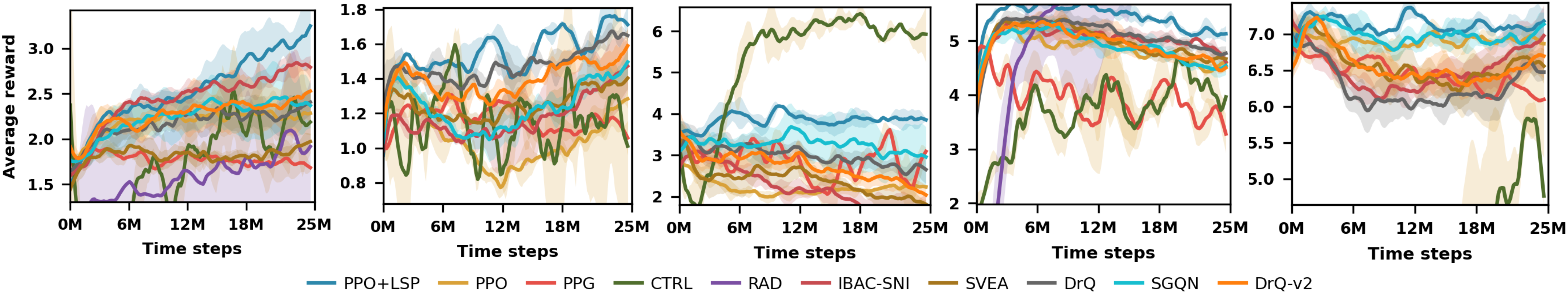}
    \vspace{-1.8em}
    \caption{
        \textbf{Procgen learning curves on Bigfish, Dodgeball, Heist, Jumper, and Plunder (left to right).}
        PPO+LSP is compared with nine baselines: PPO, PPG, CTRL, RAD, IBAC-SNI, SVEA, DrQ, SGQN, and DrQ-v2. Solid curves show average reward over training, and shaded regions indicate the variability displayed in the supplied results.
        }
    \vspace{-2.0em}
    \label{fig:procgen}
\end{figure*}

\vspace{-1.0em}
\paragraph{Concentration on top-$K$ train principal components.}
The energy ratios ($E^K_\mu, E^K_\lambda$) quantify how much OOD-induced shift is explained by dominant train directions.
For background variation, the first ten train PCs account for $87.5\%$ of the mean-shift energy and nearly all variance-shift energy.
For the other protocols, the corresponding fractions are $(0.961, 0.991)$ and $(0.852, 0.994)$.
Across all protocols, mean and variance shifts concentrate in the high-variance train subspace.
While some tail components exhibit large variance ratios (\autoref{app:metrics}), their contribution is small due to near-zero train variance.
Consistently, \autoref{fig:E_mu} shows rapid saturation of cumulative shift energy with $K$, suggesting that OOD effects are captured by a small number of leading components.
These conclusions rely on eval latents remaining aligned with the train subspace, which we verify next.

\vspace{-1em}
\subsection{Subspace similarity analysis}
\label{sec:subspace_sim}
\vspace{-1.0em}
We assess whether train and eval latents share geometry by comparing their PCA subspaces.

\vspace{-1em}
\paragraph{Train and eval PCA bases.}
Applying PCA \emph{separately} to train and eval latents yields orthonormal bases:
$V_t = (v^t_1,\ldots,v^t_D),~ V_e = (v^e_1,\ldots,v^e_D),$
referred to as the \emph{train} and \emph{eval} bases. Since both lie in the same latent feature space $\mathbb{R}^D$, they can be compared directly.

\vspace{-1em}
\paragraph{Principal angles.}
Let $\mathcal{U}=\mathrm{span}(v^t_1,\ldots,v^t_r)$ and $\mathcal{W}=\mathrm{span}(v^e_1,\ldots,v^e_r)$ be the leading train and eval subspaces.
Their alignment is measured by principal angles $\{\theta_i\}_{i=1}^r$.
Values $\cos(\theta_i)\approx 1$ indicate that dominant eval variability lies within the train subspace, while large drops signal a rotated eval subspace not captured during training.

\vspace{-1em}
\paragraph{Per-component projection score.}
Principal angles capture global alignment but not which eval directions deviate or whether deviations are significant.
Let $P_t$ be the projector onto the train subspace $\mathcal{U}$.
For a normalized eval direction $v^e_j$, define its overlap as
$\mathrm{overlap}(j)=\|P_t v^e_j\|^2,$
and the variance-weighted deviation score
$\mathrm{score}(j)=\lambda^e_j\bigl(1-\mathrm{overlap}(j)\bigr),$
where $\lambda^e_j$ is the eval eigenvalue.
Large scores arise only for directions that both lie outside the train subspace and carry substantial variance; small scores indicate variability explained by existing train directions.

\vspace{-1.0em}
\paragraph{Empirical subspace stability.}
Principal angles and variance-weighted scores assess train--eval subspace alignment and distinguish reweighting from new directions.
Across all protocols, eval latents remain almost entirely within the train PCA subspace: leading components satisfy $\cos(\theta_i)\approx 1$, with noticeable deviations only in high-index components that explain negligible variance.
Eval directions with large $\lambda^e_j$ strongly overlap with train directions and have low $\mathrm{score}(j)$, indicating amplification of existing structure rather than the emergence of new axes.
Quantitatively, the top-$100$ subspaces are tightly aligned
($\cos(\theta_0)=1.0$, $\cos(\theta_{10})\approx0.9998$, $\cos(\theta_{30})\approx0.995$),
with drops only near the tail (\autoref{fig:angles}).

\vspace{-1em}
\paragraph{Implication for the train-basis analysis.}
These results confirm that analyzing eval distributions in the train basis is valid
for all three protocols, justifying the component-level interpretations in
\autoref{sec:eval_dist_anal} and motivating interventions that regularize policy
behavior along dominant train-time latent directions. We replicate the latent space
analysis from \autoref{sec:latent-space-analysis} on Procgen tasks (Bigfish,
Dodgeball, Heist, Jumper, Plunder;~\autoref{fig:procgen_additional}), observing the
same four properties: train and eval subspaces remain closely aligned with no new
dominant directions emerging \textbf{(1)}; mean and variance discrepancies concentrate
in the first few PCs \textbf{(2)}; variance allocation ordering is preserved under OOD
observations \textbf{(3)}; and the top $K{=}10$ components retain the majority of both
$E^{K}_{\mu}$ and $E^{K}_{\lambda}$ \textbf{(4)}. The same geometry persists for task-trained CNNs under naturalistic DCS video shifts and for a frozen DINOv2 ViT under mild-to-moderate shifts, with degradation appearing only under the strongest ViT shifts (see~\autoref{app:cross_architecture_geometry}).
\contribution{
\textbf{Latent Space Analysis Takeaways}
\vspace{-0.5em}
\begin{enumerate}[leftmargin=*, itemsep=0.1em]
    \item In-distribution latents are low-rank: visual variation concentrates in leading PCs (RQ1).
    \item OOD distractors mainly induce mean shifts and variance reweighting along these leading train directions, not diffuse drift (RQ2).
    \item Eval latents stay aligned with the train PCA subspace; the tested perturbations do not introduce new dominant directions (RQ3).
\end{enumerate}
}

\vspace{-1.0em}
\subsection{From latent shift geometry to an intervention}
\label{sec:design_implications}
\vspace{-1em}
Our analysis reframes ``OOD visuals break policies'' as a geometric failure with a direct training response: eval-time variability stays aligned with the dominant train PCA subspace (\autoref{sec:subspace_sim}), meaning OOD shifts move embeddings into low-density regions rather than introducing new directions. Since shift energy concentrates in a leading PCs (\autoref{fig:E_mu}) and latent variances differ substantially across components, robustness requires variance-scaled noise along those directions, not isotropic perturbation. This motivates \textbf{Latent Space Perturbation (LSP)}, introduced in \autoref{sec:lsp_method}.

\begin{wrapfigure}{r}{0.5\textwidth}
\vspace{-3.0em}
\begin{minipage}{\linewidth}
    \vspace{-1em}
    \begin{algorithm}[H]
        \caption{
            Conceptual overview of \textbf{Latent Space Perturbation (LSP)}, derived from the latent shift geometry diagnosed in~\autoref{sec:latent-space-analysis}.
            More details in~\autoref{app:training-details},~\autoref{alg:lsp_sac},~\autoref{alg:lsp_ppo} and~\autoref{alg:lsp_forward}.
        }        
        \label{alg:lsp_conceptual}
        \small
        \begin{algorithmic}[1]
        \IF{time to recompute noise directions}
        \STATE Collect encoder latents $z = f(o)$ from training observations
        \STATE Compute PCA on train latents $\Rightarrow$ principal directions $\{v_i\}$ and variances $\{\lambda_i\}$
        \STATE Sample distraction power $p_i(t) \sim \mathcal{P}$
        \STATE Scale it: $\eta_i(t) = p_i(t)\sqrt{\lambda_i}$ 
        \ENDIF
        \FOR{each model forward during training}
            \STATE Encode observation: $z \gets f(o)$
            \STATE Sample Gaussian noise $\varepsilon \sim \mathcal{N}(0, I)$
            \STATE Perturb latent: $\hat{z} = z + \sum_i \eta_i(t)\,\varepsilon_i\,v_i$
            \STATE Generate action using policy head $a \sim \text{MLP}(\hat{z})$
        \ENDFOR
        \end{algorithmic}
    \end{algorithm}
    \vspace{-2.5em}
\end{minipage}
\end{wrapfigure}

\vspace{-1.0em}
\section{Latent Space Perturbation}
\label{sec:lsp_method}
\vspace{-1.0em}

The analysis in~\autoref{sec:latent-space-analysis} identifies a specific failure mode: appearance-only OOD inputs shift embeddings mainly within the dominant train PCA subspace, but into low-density regions rarely encountered during training.
This points to a support mismatch rather than new latent directions and does not require task-relevant and nuisance factors to occupy separate PCs. LSP only requires substantial OOD-shift energy to overlap with the dominant training subspace (see~\autoref{app:disentanglement}).
\textbf{We propose Latent Space Perturbation (LSP)}, a training-time regularizer at the representation--control interface. LSP expands the effective training support by injecting structured Gaussian perturbations aligned with dominant train PCs, scaled by the corresponding train-time variances, without modifying the encoder or using pixel-level augmentation. LSP is algorithm-agnostic: it modifies only the latent input to the policy networks during training updates, and applies to both on-policy and off-policy methods (see~\autoref{alg:lsp_sac} and~\autoref{alg:lsp_ppo}).
LSP remains lightweight: throughput decreases by about $9\%$ and GPU memory increases by $11.5\%$ on LSP-Bench, while PCA recomputation remains on the order of $10$--$15$ ms up to $D=8192$ (see~\autoref{app:lsp_overhead} and~\autoref{app:pca_scalability}). This design is principled when OOD shifts are naturalistic: 
adversarial constructions that deliberately maximize subspace mismatch 
lie outside this regime and are addressed in~\autoref{sec:limitations}.

\vspace{-1em}
\subsection{Perturbation parameterization}
\label{subsec:lsp_param}
\vspace{-1.0em}
Let $z \in \mathbb{R}^D$ denote the encoder latent for an observation.
Compute PCA on a collection of train-time latents to obtain the top-$q$ principal directions
$V_q = [v_1,\ldots,v_q] \in \mathbb{R}^{D\times q}$
and their associated eigenvalues (variances)
$\lambda=(\lambda_1,\ldots,\lambda_q)$.
Define $\sigma_i \triangleq \sqrt{\lambda_i}$ and $\sigma=(\sigma_1,\ldots,\sigma_q)$.
At training step $t$, LSP samples a dimensionless \emph{distraction power} vector
$p(t)=(p_1(t),\ldots,p_q(t))$ from a scheduler $\mathcal{P}$, with $p_i(t)\in[0,p_{\max}]$ where $p_{\max} > 0$ is a hyperparameter.
We set the per-component noise amplitudes $\eta(t) \;=\; p(t)\circ \sigma \in \mathbb{R}^{q}$,
where $\circ$ denotes element-wise multiplication.
Then we sample $\varepsilon \sim \mathcal{N}(0, I_q)$ and perturb the latent by
$
\Delta z \;=\; V_q\bigl(\varepsilon \circ \eta(t)\bigr),
~
\hat{z} \;=\; z + \Delta z
$.
Thus, each perturbation magnitude satisfies $\eta_i(t)=p_i(t)\sigma_i$, which is an explicit fraction $p_i(t)$ of one standard deviation along principal direction $v_i$.

\begin{wrapfigure}[5]{r}{0.5\textwidth}
    \vspace{-3.5em}
    \centering
    \includegraphics[width=0.5\textwidth]{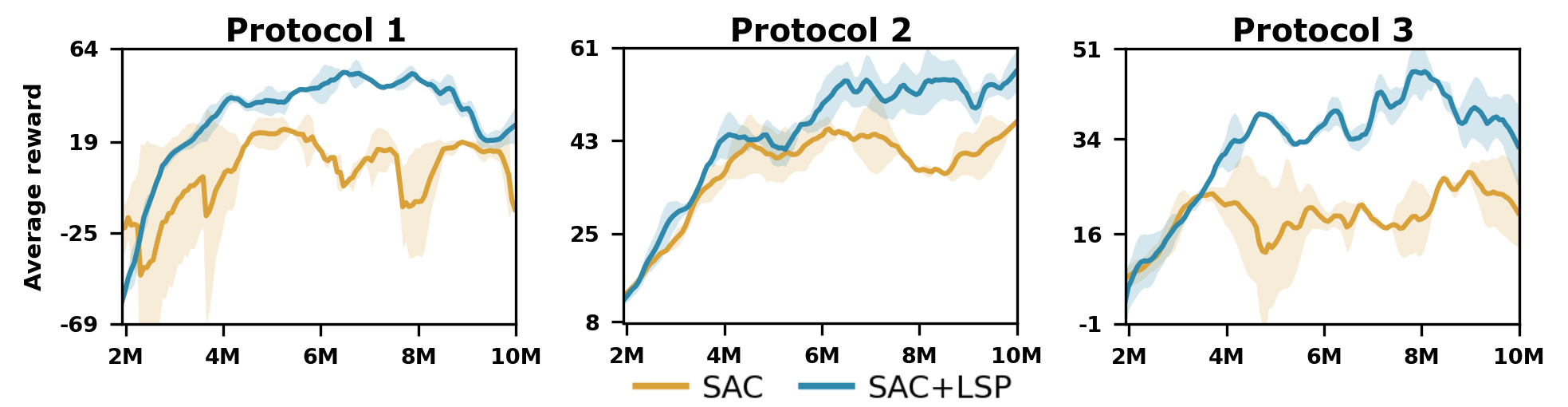}
    \vspace{-2.1em}
    \caption{
        \textbf{Performance on the LSP-Env for protocols 1, 2 and 3.}
        The avg. per-episode return increased by $20\%$ compared to PPO.
    }
    \label{fig:LSP-Env-results}
    \vspace{-2.0em}
\end{wrapfigure}

\vspace{-1em}
\subsection{Perturbation scheduler}
\label{subsec:lsp_scheduler}
\begin{wrapfigure}{r}{0.5\textwidth}
    \vspace{-2.0em}
    \centering
    \includegraphics[width=0.5\textwidth]{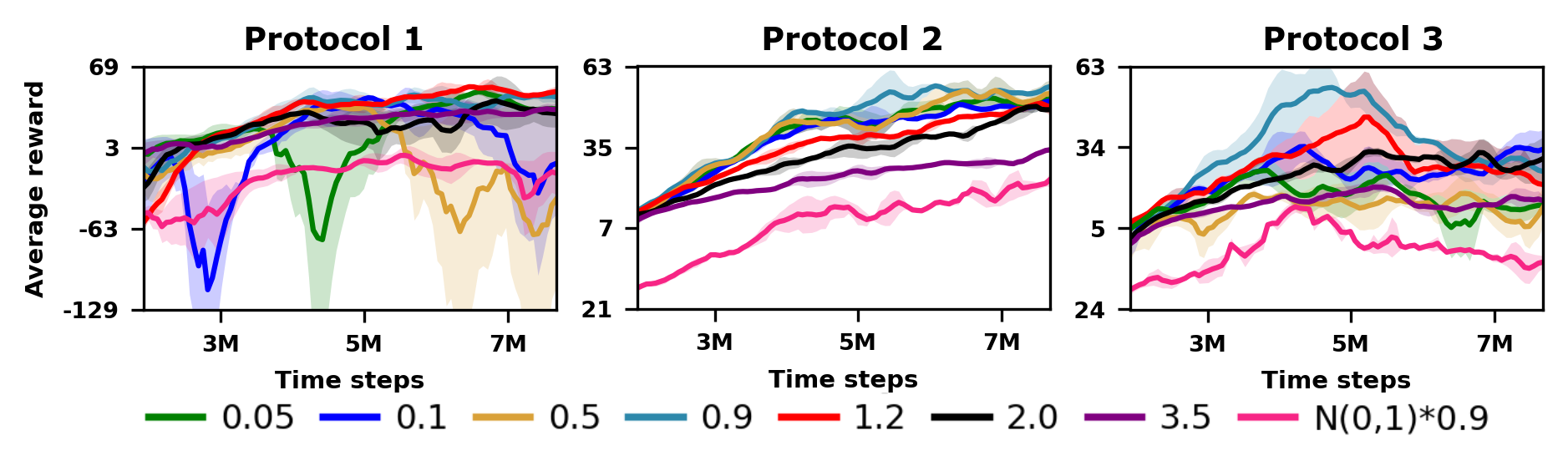}
    \vspace{-2.0em}
    \caption{
        \textbf{Ablation over noise strength.}
        Optimal performance is achieved around 0.9; weaker noise is insufficient, while stronger noise leads to lower returns.
        Uniform Gaussian noise degrades training.
    }
    \label{fig:ablation_tn}
    \vspace{-1.0em}
\end{wrapfigure}

\vspace{-1.0em}
The scheduler $\mathcal{P}$ controls the temporal profile and magnitude of perturbations through $p(t)$. We use bounded schedules (linear ramps, sudden on/off, sinusoidal, constant), detailed in~\autoref{tab:distraction_schedules}. Because $p_i(t)$ is dimensionless and $\sigma_i$ carries the latent scale, the magnitude is directly interpretable: for example, $p_i(t)=0.9$ injects noise with standard deviation $0.9\,\sigma_i$ along $v_i$.
We ablate perturbation strength (see~\autoref{fig:ablation_tn}). Varying $p_{\max}$ from $0.05$ to $3.5$ across all three protocols reveals a consistent pattern: performance peaks around $p_{\max} \approx 0.9$, with weaker noise providing insufficient regularization and stronger noise degrading performance, suggesting a moderate task-dependent optimum. At fixed perturbation magnitude, no scheduling strategy consistently dominates, while across Procgen the preferred $p_{\max}$ is more closely associated with policy sensitivity to the perturbed latent directions than with distractor count (see~\autoref{app:lsp_hyperparameter_sensitivity}). To assess whether structured perturbation is necessary at all, we compare LSP
against isotropic Gaussian noise $\mathcal{N}(0, I)$ injected into the full latent vector. Naive noise injection consistently \emph{degrades} performance relative to the unperturbed baseline, confirming that the benefit of LSP comes specifically from perturbing along the high-variance train directions, those along which OOD visual shifts concentrate.

\vspace{-1.0em}
\subsection{Results}
\label{sec:results}
\vspace{-1.0em}

\begin{wrapfigure}{r}{0.3\textwidth}
    \vspace{-1.35em}
    \centering
    \includegraphics[width=0.3\textwidth]{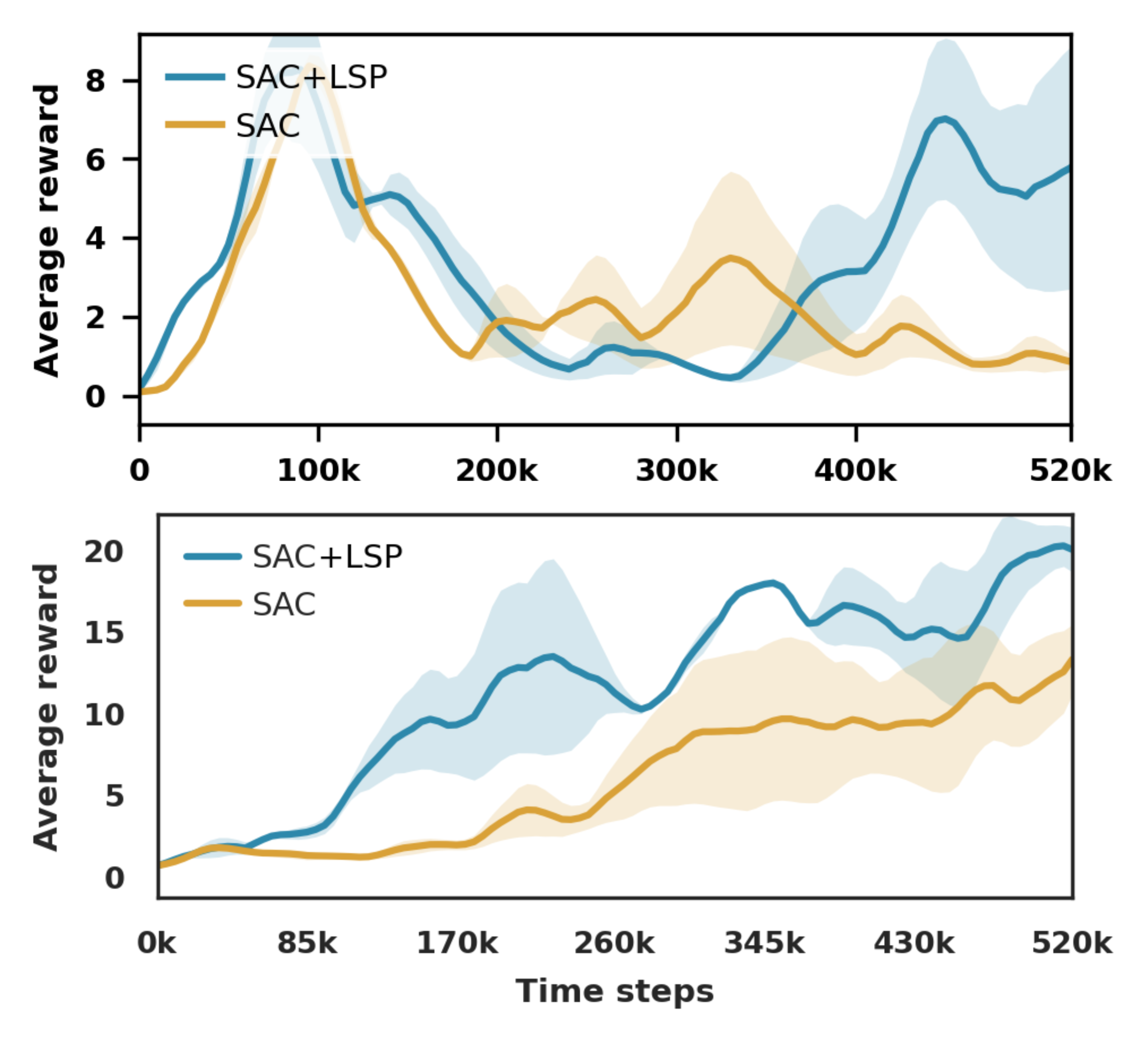}
    \vspace{-2.1em}
    \caption{
        Results on Robosuite tasks Lift (above) and Door (below) in RL-ViGen.}
    \label{fig:robosuite}
    \vspace{-1.5em}
\end{wrapfigure}

We evaluate LSP as a direct test of the analysis-to-intervention link in~\autoref{sec:latent-space-analysis} and~\autoref{sec:lsp_method}.
The diagnosis predicts that robustness should improve by training the policy head to tolerate structured displacements in the dominant train PCA subspace, without systematic in-distribution degradation since the perturbations follow directions present at train time.
Across all benchmarks, we compare each baseline agent to an LSP-enabled variant with identical architecture and optimization, differing only in whether PCA-aligned latent perturbations are applied.
We validate the mechanism under controlled appearance-only shifts in LSP-Env, and also in Procgen and RL-ViGen. For reproducibility, all training details (architectures, optimizers, hyperparameters, LSP scheduler settings, evaluation protocols, seeds, and hardware) are reported in~\autoref{app:experimental-setup}.

\vspace{-1em}
\paragraph{\textbf{Controlled appearance shifts (LSP-Env).}}

LSP-Env varies appearance while keeping dynamics fixed, so performance differences isolate visual robustness.
In \autoref{fig:LSP-Env-results}, LSP improves peak average episodic return by up to $20\%$ over PPO.
For Protocol~1 (background), the best configuration uses the sinusoidal schedule (Strategy~5) with $p_{\max}=0.9$, injecting noise up to $0.9\sigma_i$.
For Protocols~2 and~3 (agent-like distractors and agent appearance), the same schedule performs best with $p_{\max}=0.8$, reflecting the need to limit perturbations when distractors closely resemble the controlled agent.

\vspace{-1em}

\paragraph{\textbf{Uncontrolled visual diversity (Procgen).}}
We compare PPO+LSP against nine baselines: PPO~\citep{ppo}, PPG~\citep{ppg}, CTRL~\citep{ctrl}, RAD~\citep{rad}, IBAC-SNI~\citep{igl2019generalization}, SVEA~\citep{hansen2021stabilizing}, DrQ~\citep{kostrikov2020image}, SGQN, and DrQ-v2. This expanded comparison includes on-policy RL baselines, image-augmentation approaches, and representation-regularization methods. As shown in~\autoref{fig:procgen}, PPO+LSP improves over the unmodified PPO baseline in all five tasks at the end of training, with the magnitude of the advantage depending on the environment.

The largest separation from the competing curves is observed on Bigfish, where PPO+LSP continues to improve throughout training and reaches the highest displayed final reward. On Dodgeball, PPO+LSP also finishes among the strongest methods, slightly above the closest competing curves. On Heist, CTRL is the clear top performer, while PPO+LSP maintains a substantially higher final reward than vanilla PPO. On Jumper, PPO+LSP learns rapidly and remains competitive, although RAD also reaches high rewards (partly outside the displayed vertical range). On Plunder, PPO+LSP and SGQN form a leading group, with comparable performance near the end of training. These task-dependent differences show that LSP is broadly competitive without uniformly dominating every alternative.

In addition, LSP yields a smaller final visual generalization gap than PPO on all five tasks (see~\autoref{fig:gen_gap}). The best schedule--strength pairs $\{\text{strategy};p_{\max}\}$ remain: Bigfish $\{3;0.8\}$, Dodgeball $\{5;1.1\}$, Heist $\{4;0.8\}$, Jumper $\{4;1.1\}$, and Plunder $\{4;1.1\}$. A policy-sensitivity analysis of these task-dependent values is provided in~\autoref{app:lsp_hyperparameter_sensitivity}. The additional baseline comparisons are interpreted from the supplied learning curves; exact numerical rankings and significance assessments would require the underlying evaluation values.

\vspace{-1em}
\paragraph{\textbf{Continuous-control robotics (RL-ViGen).}}
On the Robosuite Door and Lift tasks, \autoref{fig:robosuite} compares SAC with and without LSP. On Door, SAC+LSP reaches an average reward near $20$ after about $520$k steps, compared with roughly $13$ for SAC. On Lift, LSP substantially improves OOD performance while the SAC baseline degrades after its early training peak. Together, these results demonstrate that the benefit of LSP extends beyond discrete-control Procgen environments to continuous-control robotic manipulation.




\vspace{-1em}
\section{Conclusion and Limitations}
\label{sec:limitations}
\vspace{-1.0em}

We study visual OOD generalization in pixel-based RL with fixed dynamics and rewards. Appearance-only shifts primarily induce structured mean and covariance drift along dominant train-time principal components rather than new latent directions. Motivated by this geometry, we propose Latent Space Perturbation (LSP), which regularizes the policy head with PCA-aligned latent noise without modifying the encoder or using pixel-level augmentation. Across LSP-Env, Procgen, and RL-ViGen, LSP improves visual OOD performance.
LSP targets naturalistic shifts in background, distractors, and appearance, where OOD variation largely remains within the training subspace. To probe its limits, we evaluate color inversion ($c \mapsto 255 - c$) and combined background, distractor, and agent-appearance shifts. Both reduce train--eval subspace alignment and introduce variation outside the dominant train basis (\autoref{fig:limitations}). LSP remains more robust than vanilla SAC, but its advantage decreases as subspace mismatch grows, marking the natural boundary of the method.

\vspace{-1em}

\bibliography{iclr2027_conference}
\bibliographystyle{iclr2027_conference}

\appendix
\section{Forms of Generalization and Visual Benchmarks}
Reinforcement learning agents encounter multiple forms of distribution shift between training and deployment. Prior work commonly distinguishes \textbf{state and layout}, \textbf{dynamics and physics}, \textbf{goal and reward}, and \textbf{visual and observation} generalization, each targeting a different source of variability.

\textbf{State and layout generalization} concerns robustness to unseen initial states or environment configurations, often addressed through procedural generation as in schema networks, MiniGrid, and Procgen~\citep{kansky2017schema,chevalier2023minigrid,procgen}. 
\textbf{Dynamics and physics generalization} focuses on changes in physical parameters such as gravity, friction, mass, or morphology, and is typically tackled using robust RL and domain randomization methods~\citep{packer2018assessing,rajeswaran2016epopt,chen2021understanding,sadeghi2016cad2rl,peng2018sim,andrychowicz2020learning,niu2024comprehensive}, as well as frameworks that explicitly vary physical properties in robotics environments, such as Robust Gymnasium~\citep{robust_gym}. 
\textbf{Goal and reward generalization} studies adaptation to unseen goals or reward specifications, with universal value function approximators, hindsight experience replay, goal-conditioned policies, and successor features providing key mechanisms~\citep{pmlr-v37-schaul15,andrychowicz2017hindsight,liu2022goal,barreto2017successor}. 
\textbf{Visual and observation generalization} addresses robustness to appearance changes, including textures, colors, backgrounds, and camera pose, using techniques such as domain randomization, natural-video distractors, and visual augmentations or regularization~\citep{procgen,zhang2018natural,rad,kostrikov2020image,igl2019generalization}.

Several benchmarks explicitly target visual generalization. Procgen~\citep{procgen} and its configurable extension C-Procgen~\citep{c_procgen} provide procedurally generated arcade-style environments with held-out levels for evaluating generalization under visual and layout variation. The Distracting Control Suite (DCS)~\citep{dcs}, built on the DeepMind Control Suite~\citep{dms}, evaluates robustness to changes in background, color, and camera pose. More recently,~\citet{liu2025can} introduced a VLA generalization benchmark based on language-conditioned tasks, evaluating out-of-distribution robustness along visual, semantic, and execution axes.

\section{LSP-Env Additional Details}
\label{app:LSP-Env-details}

LSP-Env is a lightweight, side-scrolling 2D platformer designed for fast, controlled studies of visual generalization. 
Compared to Procgen, which offers broad procedural diversity but limited factor isolation, LSP-Env allows explicit, manual control of individual appearance factors while keeping the underlying dynamics and reward computation fixed. 
This enables rapid ablations (for example, changing only the background while keeping the same task logic) and mechanistic diagnosis of appearance-induced distribution shift.

\paragraph{Task and episode structure.}
At each step, the agent aims to move right as efficiently as possible in a procedurally generated terrain containing obstacles (gaps and vertical walls). 
The environment is side-scrolling: the agent is rendered near the horizontal center of the frame, while the world coordinate $\texttt{world\_position}$ advances as the agent moves right.
Episodes terminate only due to a step-limit truncation ($\texttt{max\_steps}$).

\paragraph{Observation space.}
Observations are RGB images of size $64 \times 64$ rendered from the current view:
\begin{equation}
\mathcal{O} = \texttt{Box}(0,255,(3,64,64),\texttt{uint8}).
\end{equation}

\paragraph{Action space.}
Actions are continuous and two-dimensional,
\begin{equation}
\mathcal{A} = \texttt{Box}\big( ( -1, 0 ), ( 1, 1 ), (2,), \texttt{float32} \big),
\end{equation}
where the first coordinate controls horizontal motion ($a_x \in [-1,1]$) and the second coordinate is a jump command ($a_j \in [0,1]$). 
The jump command is only applied when the agent is on the ground and a minimum jump interval has elapsed; in that case, it sets the initial upward velocity proportional to $a_j$.

\paragraph{Reward.}
The per-step reward is shaped to favor fast horizontal progress and discourage unnecessary actions:
\begin{equation}
r_t = \texttt{step\_reward}\cdot v_x + \texttt{step\_penalty} + \mathbb{I}[\texttt{jumped}]\cdot \texttt{jump\_penalty},
\end{equation}
where $v_x$ is the executed horizontal velocity for that step. By default, $\texttt{step\_reward}=1.0$, $\texttt{step\_penalty}=-0.1$, and $\texttt{jump\_penalty}=-0.3$.

\paragraph{Controllable axes of visual variability.}
We expose three sources of task-irrelevant appearance variation that are used throughout the paper:
\begin{itemize}
    \item \textbf{Background variation.}
    The background is sampled independently of the task state (for example: solid colors, noise, gradients, procedural textures, or natural images), enabling background-only distribution shifts.

    \item \textbf{Agent-like distractors.}
    Non-controllable geometric entities are rendered as moving distractors (with configurable number, shape, color, and optional rotation). They affect the pixels but do not affect the physical state or reward computation.

    \item \textbf{Agent appearance.}
    The controlled agent's rendering is varied across episodes (implemented via configurable shape and color, and optional rotation), without changing dynamics or the reward function.
\end{itemize}

\section{Architecture Details}
\label{app:arch-details}

\paragraph{CNN Encoder.}
We use the standard Nature-CNN backbone to map an RGB observation
$o \in \{0,\ldots,255\}^{C\times H\times W}$ to a latent embedding $z\in\mathbb{R}^{D}$.
We first normalize $x=o/255$ and compute
\begin{equation}
h_{\mathrm{conv}}=\mathrm{Flatten}\!\left(
\phi\!\left(\mathrm{Conv}_{64,3\times 3,s=1}\!\left(
\phi\!\left(\mathrm{Conv}_{64,4\times 4,s=2}\!\left(
\phi\!\left(\mathrm{Conv}_{32,8\times 8,s=4}(x)\right)\right)\right)\right)\right)\right),
\end{equation}
where $\phi(\cdot)=\mathrm{ReLU}(\cdot)$. Let $d_{\mathrm{conv}}=\dim(h_{\mathrm{conv}})$. We then form a normalized hidden representation
\begin{equation}
h=\mathrm{LN}\!\left(\phi\!\left(W h_{\mathrm{conv}} + b\right)\right)\in\mathbb{R}^{D},
\qquad D=256,
\end{equation}
and output the final latent embedding via a linear projection
\begin{equation}
z = W_{\mathrm{out}} h + b_{\mathrm{out}} \in \mathbb{R}^{D}.
\end{equation}

\begin{table*}[t]
\centering
\small
\setlength{\tabcolsep}{6pt}
\caption{Metrics used to characterize train-eval latent distribution shift, all computed in the train PCA basis.}
\begin{tabular}{lll}
\toprule
\textbf{Metric} & \textbf{Definition} & \textbf{Interpretation} \\
\midrule
$\text{mean\_diff}_i$
& $|\mu_i^{e} - \mu_i^{t}|$
& Absolute mean shift along the $i$-th train PC \\

$\text{var\_diff}_i$
& $|\lambda_i^{e} - \lambda_i^{t}|$
& Absolute change in variance along the $i$-th train PC \\

$\text{var\_ratio}_i$
& $\lambda_i^{e} / \lambda_i^{t}$
& Relative variance change along the $i$-th train PC \\

$\text{var\_frac}^{t/e}_i$
& $\lambda_i^{t/e} \big/ \sum_{j=1}^{D} \lambda_j^{t/e}$
& Fraction of total variance explained by the $i$-th train PC \\

$\text{frac\_ratio}_i$
& $\text{var\_frac}^{e}_i \big/ \text{var\_frac}^{t}_i$
& Change in explained variance between eval and train data of the $i$-th PC \\

$E_{\mu}^{K}$
& $\frac{1}{E_\mu}\sum_{i \in \mathrm{Top\text{-}}K(\lambda^t)} (\Delta \mu_i)^2$
& Mean-shift energy ratio in top-$K$ train PCs \\

$E_{\lambda}^{K}$
& $\frac{1}{E_\lambda}\sum_{i \in \mathrm{Top\text{-}}K(\lambda^t)} (\Delta \lambda_i)^2$
& Variance-shift energy ratio in top-$K$ train PCs \\
\bottomrule
\end{tabular}
\label{tab:latent_shift_metrics}
\end{table*}

\newpage
\section{Experimental Setup}
\label{app:experimental-setup}

This appendix reports the full configuration used to produce the results in~\autoref{sec:results}, covering all three benchmarks (LSP-Env, Procgen, RL-ViGen). We list the model architecture, optimizer, RL-algorithm hyperparameters, LSP-specific scheduler settings, evaluation protocol, and compute hardware. Hyperparameter values not specified explicitly follow the reference implementations of SAC and PPO.

\subsection{Hardware and Software}
All experiments were run on a single workstation: NVIDIA RTX 3060 (12\,GB VRAM), Intel Core i7-10700K, 32\,GB RAM. Training used PyTorch with mixed-precision disabled, Adam optimizers, and TensorBoard logging. PCA recomputation for LSP was performed on GPU using torch.

\subsection{Common Components}
All three benchmarks share the same conceptual pipeline: a CNN encoder produces a latent $z \in \mathbb{R}^D$ from RGB observations, an MLP policy head maps $z$ to actions, and LSP injects PCA-aligned Gaussian noise into $z$ during training. The PCA basis is recomputed every $f$ steps from a buffer of recent encoder latents and used until the next refresh. The noise scheduler is identical across benchmarks: a strategy index in $\{1,\dots,6\}$ (\autoref{tab:distraction_schedules}) parameterized by $p_{\max}$ (target noise), $t_{\text{switch}}$, and the number of cycles.

\subsection{Per-Benchmark Configuration}
\autoref{tab:setup-master} consolidates the full configuration for each benchmark. A few items are worth highlighting before reading the table:
\begin{itemize}[leftmargin=*, itemsep=0em]
    \item LSP-Env and RL-ViGen use SAC; Procgen uses PPO. Scheduler hyperparameters reported below correspond to the best configurations identified in~\autoref{sec:results}.
    \item In LSP-Env the encoder follows the Nature-CNN backbone. RL-ViGen uses a deeper $3\times 3$ stride-2 stack to handle $84\times 84$ inputs. Procgen uses the same Nature-CNN output dimension as in~\citet{procgen}.
    \item The PCA recompute period $f$ (\texttt{mainvec\_frequency}) is given in environment steps for SAC and in policy iterations for PPO.
\end{itemize}

\begin{table}[h]
\caption{Full experimental configuration across the three benchmarks. Values not listed follow reference SAC/PPO implementations.}
\label{tab:setup-master}
\centering
\small
\begin{tabular}{l l l l}
\toprule
& \textbf{LSP-Env} & \textbf{Procgen} & \textbf{RL-ViGen (Door)} \\
\midrule
\multicolumn{4}{l}{\textit{Algorithm and observations}} \\
RL algorithm                & SAC                & PPO                & SAC \\
Observation size            & $64\times 64\times 3$ & $64\times 64\times 3$ & $84\times 84\times 3$ \\
Action space                & continuous, 2D     & discrete (15)      & continuous, 4D \\
Frame stack                 & 1                  & 1                  & 1 \\
Total env steps             & $1.0\times 10^7$   & $2.5\times 10^7$   & $5.0\times 10^5$ \\
\midrule
\multicolumn{4}{l}{\textit{Encoder (CNN $\to$ latent $z\in\mathbb{R}^D$)}} \\
Conv stack                  & Nature-CNN ($32{-}64{-}64$) & IMPALA / Nature-CNN & $32{-}32{-}32$, $3\times 3$ s2 $\times 3$ \\
Latent dim. $D$             & 256                & 256                & 1024 \\
Post-conv MLP               & FC$\to$LN$\to$FC   & FC$\to$LN$\to$FC   & FC$\to$LN$\to$FC \\
\midrule
\multicolumn{4}{l}{\textit{Policy / value heads}} \\
Policy MLP                  & $[256,256]$        & $[256,64]$         & $[512,512]$ \\
Q / value MLP               & $[256,256]$        & shared trunk       & $[512,512]$ \\
\midrule
\multicolumn{4}{l}{\textit{Optimization}} \\
Optimizer                   & Adam               & Adam ($\epsilon{=}10^{-5}$) & Adam \\
Policy LR                   & $1\times 10^{-4}$  & $2.5\times 10^{-4}$ & $1\times 10^{-4}$ \\
Critic / value LR           & $3\times 10^{-4}$  & shared              & $1\times 10^{-4}$ \\
LR schedule                 & constant           & linear decay        & constant \\
Grad-clip norm              & $10.0$             & $0.5$               & $10.0$ \\
Discount $\gamma$           & $0.95$             & $0.99$              & $0.99$ \\
GAE $\lambda$               & --                 & $0.95$              & -- \\
Entropy coef.               & autotuned ($\alpha$) & $0.01$            & autotuned ($\alpha$) \\
Value coef.                 & --                 & $0.5$               & -- \\
PPO clip $\epsilon$         & --                 & $0.2$               & -- \\
\midrule
\multicolumn{4}{l}{\textit{SAC / PPO specific}} \\
Parallel envs               & 128                & 8                   & 1 \\
Rollout length / batch      & $b{=}512$          & $n_{\text{steps}}{=}128$ & $b{=}512$ \\
PPO epochs / minibatches    & --                 & $3$ / $8$           & -- \\
Replay buffer size          & $5\times 10^5$     & --                  & $4\times 10^5$ \\
Learning starts             & 500                & --                  & $10\,000$ \\
Target update $\tau$        & $0.005$            & --                  & $0.005$ \\
Encoder target $\tau_{\text{enc}}$ & $0.001$     & --                  & $0.001$ \\
Policy update freq.         & every 2 steps      & --                  & every 2 steps \\
\midrule
\multicolumn{4}{l}{\textit{LSP-specific}} \\
Perturbed directions $q$    & $4$                & $1$                 & $1$ \\
PCA recompute period $f$    & 250 env steps      & every 1 PPO iter.   & 5\,000 env steps \\
Noise strategy & 5 (sinusoidal) & 3 (sudden on) & 3 (sudden on) \\
Target noise $p_{\max}$     & 0.9 (P1) / 0.8 (P2,P3) & task-dependent (\autoref{sec:results}) & 0.8 \\
$t_{\text{switch}}$         & $5\times 10^6$     & $10^7$              & $5\times 10^6$ \\
Cycles                      & 1                  & 4                   & 1 \\
\midrule
\multicolumn{4}{l}{\textit{Evaluation}} \\
Eval frequency              & every 500 env steps & every PPO iter. on held-out levels & every $5\,000$ env steps \\
Eval episodes / envs        & 100 parallel       & 128 (start\_level=0, num\_levels=0) & 10 \\
Train / eval split          & fixed visual config & 200 train levels / unbounded eval & original \texttt{train} / \texttt{eval} split \\
\bottomrule
\end{tabular}
\end{table}

\subsection{Seeds, Aggregation, and Error Bars}
Each configuration is trained with independent random seeds (seeds 1--5), differing in environment sampling, network initialization, and replay buffer order. Learning curves report the mean across seeds; shaded regions correspond to standard deviation across seeds after applying a moving average over a window of $10$ consecutive evaluation points to reduce point-wise noise.

\section{Latent Space Analysis Metrics}
\label{app:metrics}

\paragraph{\textbf{Mean shift (\texttt{mean\_diff}).}}
The metric $\text{mean\_diff}_k = |\mu_k^e-\mu_k^t|$ measures the absolute displacement of the eval mean along the $k$-th train principal component. As shown in \autoref{fig:vis_diff}, the largest mean shifts occur along leading components ($k=0,1,2,3,7,\dots$) with $\text{mean\_diff}\in[2.9,20]$, and these components also have high train variance $\lambda_k^t$. Thus, background-induced OOD perturbations primarily translate the latent distribution along the dominant (high-variance) train directions rather than along the low-variance tail. Since all task-relevant factors are held fixed in the eval construction, these shifts can be attributed to background variation.

\paragraph{\textbf{Train variance concentration (\texttt{var\_frac}$^t$).}}
The metric $\text{var\_frac}_k^t = \lambda_k^t/\sum_j \lambda_j^t$ captures the fraction of total train variance explained by component $k$. The train latent space is highly concentrated: $v_0$ explains approximately $60\%$ of the variance and the top ten components account for more than $0.88$. This indicates that the dominant structure of the train distribution is low-dimensional.

\paragraph{\textbf{Eval variance in the train basis (\texttt{var\_frac}$^e$).}}
Projecting eval latents onto the train basis, $\text{var\_frac}_k^e = \lambda_k^e/\sum_j \lambda_j^e$ shows that eval variance is likewise concentrated in the leading train directions (e.g., $k=1,0,7,3,5,2,\dots$). Moreover, analyzing the variance-change energy $\sum_k (\Delta\lambda_k)^2$ reveals that variance changes are almost entirely confined to directions that are already dominant during training: the top five PCs by $\lambda_{\text{train}}$ account for approximately $0.998$ of $\sum_k (\Delta\lambda_k)^2$ (\autoref{fig:vis_diff}). This supports the hypothesis that OOD variability largely reweights variance within the principal train subspace.

\paragraph{\textbf{Relative importance shift (\texttt{frac\_ratio}).}}
The metric $\text{frac\_ratio}_k = \text{var\_frac}_k^e/\text{var\_frac}_k^t$ measures how the \emph{relative} contribution of each component to total variance changes from train to eval. For intermediate components with large $\text{var\_diff}$ (e.g., $k=27,34,42,47,\dots$), $\text{frac\_ratio}_k$ can be substantial (approximately $6.9$--$11$), indicating that these directions become relatively more prominent under eval. However, their absolute explained variance remains small, since both $\text{var\_frac}^t$ and $\text{var\_frac}^e$ are on the order of $10^{-4}$--$10^{-2}$. Conversely, components with the largest $\text{var\_frac}^e$ (e.g., $k=0,1,2,3,\dots$) typically exhibit $\text{frac\_ratio}_k$ in the range $0.3$--$2.2$, indicating modest redistribution among the leading directions. Extremely large $\text{frac\_ratio}_k$ values in the far tail (where $\text{var\_frac}^t \sim 10^{-6}$ and $\text{var\_frac}^e \sim 10^{-4}$) are not indicative of meaningful structure, as their absolute contributions remain negligible.

\paragraph{\textbf{Relative variance change (\texttt{var\_ratio}).}}
The metric $\text{var\_ratio}_k = \lambda_k^e/\lambda_k^t$ characterizes multiplicative variance change. Extremely large ratios (on the order of $10^3$ and above) occur only when $\lambda_k^t$ is near zero (e.g., $\lambda_k^t\sim10^{-13}$, $\text{var\_frac}_k^t\sim10^{-16}$), and these components also have negligible eval variance ($\text{var\_frac}_k^e\sim10^{-7}$). Such values are therefore dominated by denominator effects rather than meaningful variance growth. For components with non-negligible train variance, $\text{var\_ratio}_k$ remains moderate (e.g., $1.4$--$2.1$), reflecting genuine increases in variability.

\paragraph{\textbf{Absolute variance change (\texttt{var\_diff}).}}
The metric $\text{var\_diff}_k = |\lambda_k^e-\lambda_k^t|$ identifies directions with the largest absolute increase in spread and is not distorted by small denominators. The largest values ($\sim 0.3$--$0.8$) occur at intermediate components ($k=27,34,42,47,\dots$), where $\lambda_k^t\approx0.5$--$1.2$, $\text{var\_frac}_k^t\sim10^{-3}$--$10^{-4}$, and $\text{var\_frac}_k^e\sim10^{-2}$--$10^{-3}$. Thus, although these directions exhibit noticeable absolute variance increases, they account for a small fraction of total variance both before and after the shift. Finally, because the eval construction isolates background variability while freezing other factors, eval variance is likely underestimated relative to settings where all sources of variation are active. Nevertheless, even under this constrained setup, certain leading components (e.g., PC1, PC5, PC7) exhibit increased variance driven by background variation alone.

\begin{figure}[b]
    \centering
    \includegraphics[width=1\linewidth]{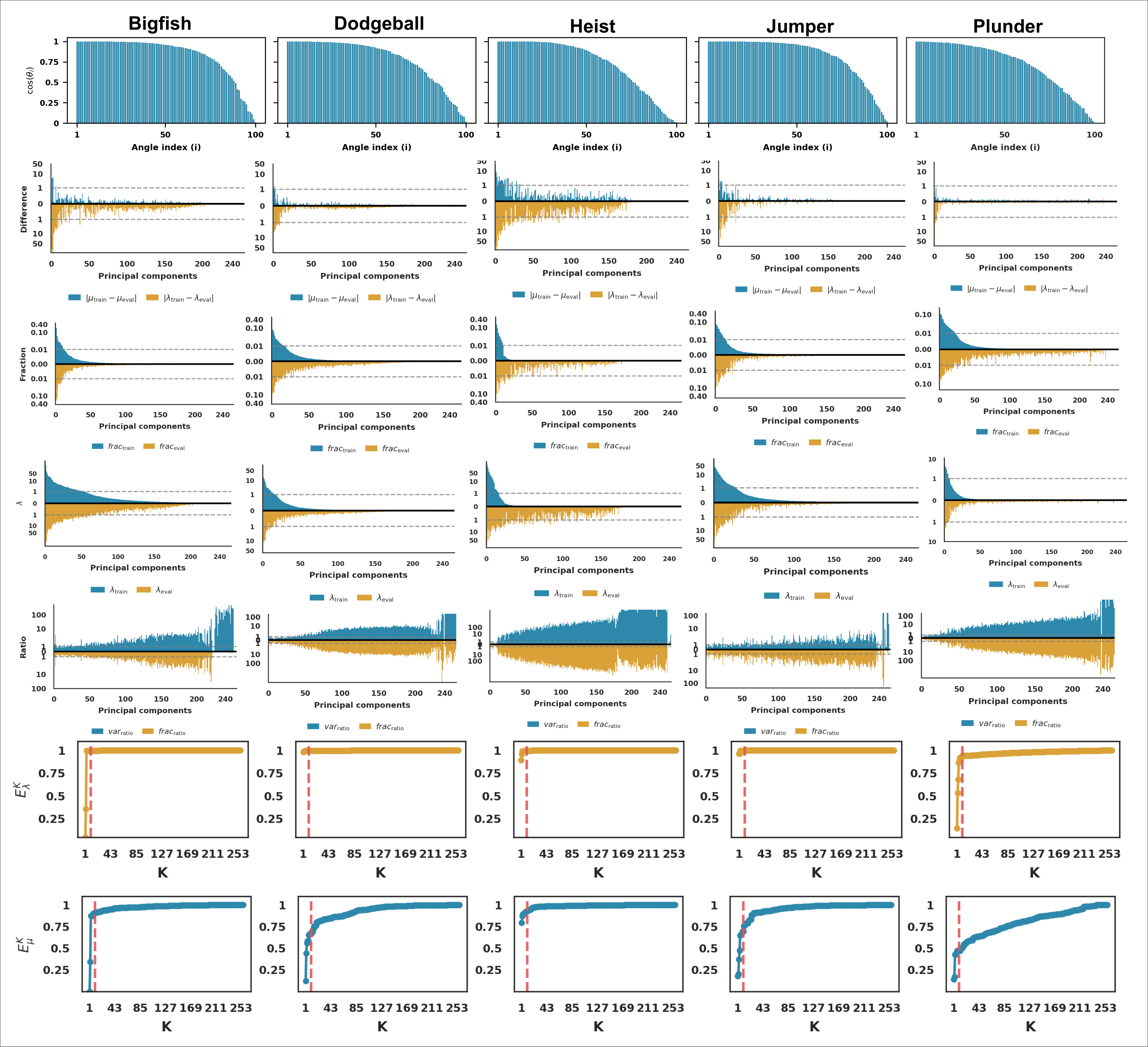}
    \caption{\textbf{Latent space analysis on Procgen environments.} Results demonstrate the same properties as on LSP-Env: \textbf{1)} Train and Eval spaces are similar to each other and do not introduce new variance directions (line 1)
    \textbf{2)}The biggest differences between $\mu$ and $\lambda$ occurs along the first principal components (line 2),
    \textbf{3)} During evaluation OOD observations processing, variation allocation follows the same pattern as in tran ID data (line 4)
    \textbf{4)} The first K (K=10) components still preserve the most of the mean-shift energy $E^{K}_{\mu}$ and mean-variance energy $E^{K}_{\lambda}$ (lines 6,7).}
    \label{fig:procgen_additional}
\end{figure}

\begin{figure}[t]
    \centering

    \begin{subfigure}{\linewidth}
        \centering
        \includegraphics[width=1\linewidth]{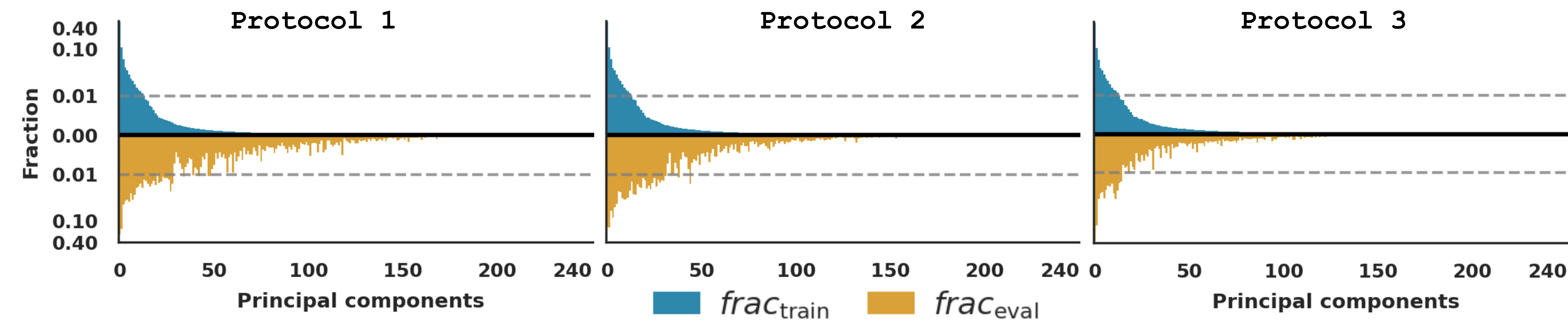}
        \caption{Fraction of training and eval explained variance within each principal component in train basis.}
        \label{fig:frac_te}
    \end{subfigure}

    \vspace{0.4em}

    \begin{subfigure}{\linewidth}
        \centering
        \includegraphics[width=1\linewidth]{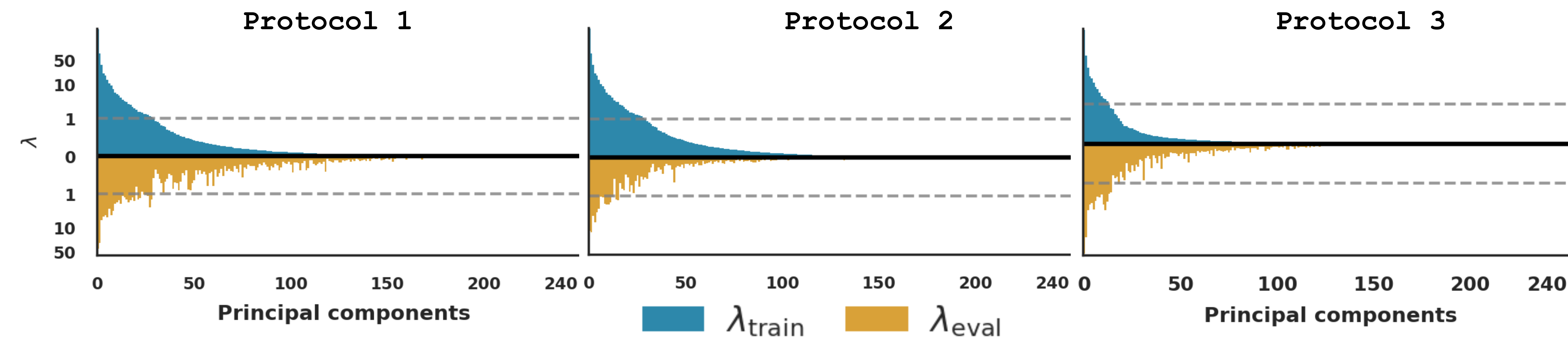}
        \caption{Training and eval variances within each principal component in train basis.}
        \label{fig:lambda_trash}
    \end{subfigure}

    \vspace{0.4em}

    \begin{subfigure}{\linewidth}
        \centering
        \includegraphics[width=1\linewidth]{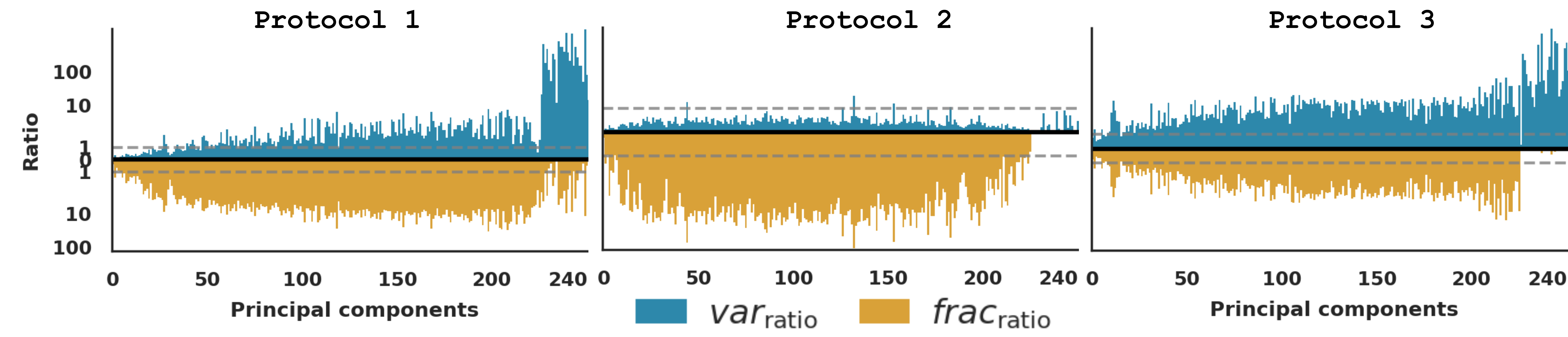}
        \caption{Variance ratios for train and eval data in the train basis.}
        \label{fig:ratios}
    \end{subfigure}

    \label{fig:train_eval_variance_panels}
\end{figure}

\begin{figure}[b]
    \centering
    \includegraphics[width=0.9\linewidth]{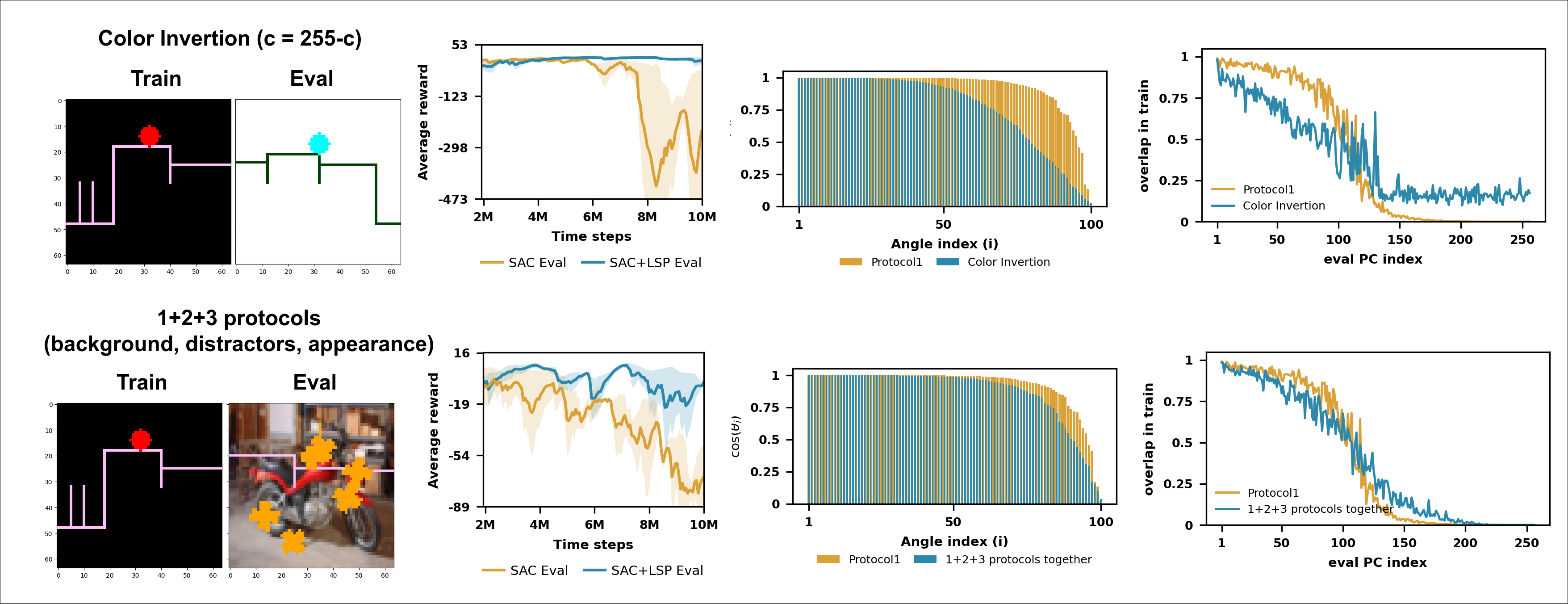}
    \caption{In the figures, we present sample observations, performance comparisons between SAC and SAC+LSP, and the behavior of \textbf{subspace-angle cosines} and the \textbf{overlap} metric relative to the background-only setting. This comparison is included because background changes, while visually salient (change $\sim$ 90\% of the observation), do not introduce new directions of variability according to our analysis. The \textbf{overlap} metric is defined as the squared norm of the projection of a normalized eval direction onto the train subspace, i.e., $\mathrm{overlap}(j)=|P_t v_j^e|^2$, and thus measures how well a given eval direction is explained by the dominant train subspace.
We emphasize that subspace-angle cosines capture changes in variability directions, whereas the overlap metric indicates whether dominant variability remains concentrated in leading coordinates of the train basis.}
    \label{fig:limitations}
\end{figure}

\section{Scheduling}
\label{app:scheduling}
LSP controls perturbation magnitude over training with a \textit{distraction power} scheduler that modulates the per-step noise scale.
This is useful for two reasons: (i) it avoids imposing the same perturbation strength throughout optimization, which can be overly disruptive early on or unnecessarily mild later; and (ii) it lets us probe how robustness depends on the timing and persistence of latent perturbations.

Concretely, at step $t$ the scheduler outputs a bounded scalar (or vector) $p(t)\in[0,p_{\max}]$ that determines the perturbation amplitude applied in the train PCA basis (see~\autoref{subsec:lsp_param}).
Because $p(t)$ is dimensionless, $p_{\max}$ is directly interpretable as the maximum fraction of a train-time standard deviation used for perturbation along selected components.
We consider schedules that vary in \textit{onset} (gradual vs.\ abrupt), \textit{duration} (transient vs.\ persistent), and \textit{periodicity} (constant vs.\ oscillatory), enabling controlled ablations of when and how strongly the policy head is regularized.
\autoref{tab:distraction_schedules} lists the schedules used in our experiments.

\begin{table*}[t!]
\centering
\small
\setlength{\tabcolsep}{6pt}
\caption{Distraction power scheduling strategies used in LSP.}
\begin{tabular}{lll}
\toprule
\textbf{Schedule} & \textbf{Power function $p(t)$} & \textbf{Description} \\
\midrule
Linear ramp-up
& $p_{\max}\,\dfrac{t}{T}$
& Linearly increases distraction from $0$ to $p_{\max}$ \\

Linear ramp-down
& $p_{\max}\!\left(1 - \dfrac{t}{T}\right)$
& Linearly decreases distraction from $p_{\max}$ to $0$ \\

Sudden activation
& $\begin{cases}
0, & t < t_{\text{switch}},\\
p_{\max}, & t \ge t_{\text{switch}}
\end{cases}$
& Abruptly turns distraction on at $t_{\text{switch}}$ \\

Sudden deactivation
& $\begin{cases}
p_{\max}, & t < t_{\text{switch}},\\
0, & t \ge t_{\text{switch}}
\end{cases}$
& Abruptly turns distraction off at $t_{\text{switch}}$ \\

Cyclical sinusoidal
& $\dfrac{p_{\max}}{2}\!\left[1 + \sin\!\left(2\pi c\,\dfrac{t}{T} - \dfrac{\pi}{2}\right)\right]$
& Periodic modulation with frequency $c$ \\

Constant
& $p_{\max}$
& Fixed distraction strength throughout training \\
\bottomrule
\end{tabular}
\label{tab:distraction_schedules}
\end{table*}

\section{Algorithm and Training Details}
\label{app:training-details}
This section summarizes the training configurations used in our experiments. We report the key hyperparameters and model architectures for both PPO, used in Procgen, and SAC, used in LSP-Env. Unless stated otherwise, all other settings follow standard implementations, and the reported values were selected based on preliminary sweeps and prior work to ensure stable training and fair comparison across methods.

\begin{table}[h!]
\centering
\begin{minipage}{0.48\textwidth}
\centering
\caption{\textbf{PPO hyperparameters and model size (Procgen).}}
\label{tab:ppo}
\vspace{2pt}
\begin{tabular}{l c}
\toprule
\textbf{Parameter} & \textbf{Value} \\
\midrule
Number of environments ($n_{\text{envs}}$) & 8 \\
Epochs & 3 \\
Mini-batch size & 256 \\
Discount factor $\gamma$ & 0.99 \\
GAE parameter $\lambda$ & 0.95 \\
Learning rate & $2.5\times 10^{-4}$ \\
Gradient clipping norm & 0.5 \\
Entropy coefficient & 0.01 \\
Value loss coefficient & 0.5 \\
\midrule
\textbf{Model size} & \\
CNN output dimension & 256 \\
MLP hidden dimensions & [256, 64] \\
\bottomrule
\end{tabular}
\end{minipage}
\hfill
\begin{minipage}{0.48\textwidth}
\centering
\caption{\textbf{SAC hyperparameters and model size (LSP-Env).}}
\label{tab:sac}
\vspace{2pt}
\begin{tabular}{l c}
\toprule
\textbf{Parameter} & \textbf{Value} \\
\midrule
Total timesteps & 10,000,000 \\
Learning starts & 500 \\
Replay buffer size & 500,000 \\
Batch size & 512 \\
Discount factor $\gamma$ & 0.95 \\
Target update rate $\tau$ & 0.005 \\
Policy update frequency & 2 \\
Q-network learning rate & $3\times 10^{-4}$ \\
Policy learning rate & $1\times 10^{-4}$ \\
Entropy coefficient $\alpha$ & 0.2 \\
\midrule
\textbf{Model size} & \\
CNN output dimension & 256 \\
Policy MLP layers & [256, 256, $\text{act\_dim}$] \\
Q-network MLP layers & [256, 256, 1] \\
\bottomrule
\end{tabular}
\end{minipage}
\end{table}

\newpage
\clearpage
\begin{algorithm}[t]
    \caption{SAC with Latent Space Perturbation. \textcolor{myblue}{Highlighted components indicate modifications introduced by LSP relative to standard SAC.}}
    \label{alg:lsp_sac}
    \small
    \begin{algorithmic}[1]
       \INPUT policy $\pi_\theta$, critics $Q_{\theta_1},Q_{\theta_2}$, \lsp{perturbation scheduler $\mathcal{P}$, PCA period $f$, number of components $q$}
       \SET replay buffer $\mathcal{B} \leftarrow \emptyset$
       \STATE Initialize target networks $\theta' \leftarrow \theta$, $\theta'_1 \leftarrow \theta_1$, $\theta'_2 \leftarrow \theta_2$
       \FOR{$t \gets 1,\ldots,T$}
       \STATE Sample $a \sim \pi_\theta(o \mid s)$
       \STATE Execute $a$, observe $r$, $s'$, and done flag $d$
       \STATE Store transition $(s, a, r, s', d)$ in $\mathcal{B}$
       \lspstart
       \IF{$t \bmod f = 0$}
          \STATE Recompute PCA basis from samples in $\mathcal{B}$:
          \STATE $\mathbf{V} \gets [v_1,\ldots,v_q]$, $\mathbf{e} \gets (e_1,\ldots,e_q)$ \COMMENT{top-$q$ principal directions and their variances}
       \ENDIF
       \lspend
       \STATE Sample batch $(s, a, r, s',d)$ from $\mathcal{B}$
       \STATE Sample $\tilde a \sim \pi_{\theta}(o \mid s')$ and set $\mathcal{H} \gets -\log \pi_{\theta}(\tilde a \mid s')$
       \STATE $y \gets r+\gamma(1 - d)\big(\min_i\{Q_{\theta'_{i}}(s',\tilde a)\}+\alpha \mathcal{H}\big)$
       \STATE Update critics: $\theta_i \leftarrow \arg\min_{\theta_i} N^{-1} \sum \big(y - Q_{\theta_i}(s,a)\big)^2$
       \STATE {\color{lspcolor}Sample perturbation $\mathbf{p} = (p_1,\ldots,p_q) \sim \mathcal{P}(t)$}
       \STATE {\color{lspcolor}Set $\mathbf{z} \gets \mathbf{p} \circ \sqrt{\mathbf{e}}= (p_1\sqrt{e_1},\ldots,p_q\sqrt{e_q})$}
       \STATE {\color{lspcolor}Sample $\hat a_{\theta} \sim \pi_{\theta}(o \mid s,\mathbf{z},\mathbf{V})$ via \autoref{alg:lsp_forward}}
       \STATE {\color{lspcolor}Set $\mathcal{H} \gets -\log \pi_{\theta}(\hat a_{\theta} \mid s)$}
       \STATE {\color{lspcolor}Update actor: $\nabla_{\theta} N^{-1} \sum \big(\min_i\{Q_{\theta_{i}}(s,\hat a_{\theta})\}+\alpha \mathcal{H}\big)$}
       \STATE Update targets: $\theta'_i \leftarrow \tau \theta_i + (1 - \tau) \theta'_i,\;\theta' \leftarrow \tau \theta + (1 - \tau) \theta'$
       \ENDFOR
    \end{algorithmic}
    \end{algorithm}

\begin{minipage}{\linewidth}
    \vspace{-1.5em}
    \begin{algorithm}[H]
        \caption{PPO with Latent Space Perturbation (LSP). \textcolor{myblue}{Highlighted components indicate modifications introduced by LSP relative to standard PPO.}}
        \label{alg:lsp_ppo}
        \small
        \begin{algorithmic}[1]\raggedright
           \INPUT policy $\pi_\theta$, value $V_\psi$, \lsp{perturbation scheduler $\mathcal{P}$, PCA period $f$, number of components $q$}, clip $\epsilon$, epochs $E$
           \STATE Initialize policy/value parameters $\theta,\psi$
           \FOR{$k \gets 1,\ldots,K$ \COMMENT{policy iterations}}
              \STATE Collect on-policy rollout $\mathcal{D}$ of length $T$ using $\pi_{\theta_k}$
              \STATE Compute returns $\hat{R}_t$ and advantages $\hat{A}_t$ from $\mathcal{D}$
              \lspstart
              \IF{$k \bmod f = 0$}
                 \STATE Recompute PCA on encoder latents in $\mathcal{D}$
                 \STATE $\mathbf{V} \gets [v_1,\ldots,v_q]$, $\mathbf{e} \gets (e_1,\ldots,e_q)$ \COMMENT{top-$q$ PCs}
              \ENDIF
              \lspend
        
              \FOR{$\text{epoch} \gets 1,\ldots,E$}
                 \STATE Sample minibatch $\mathcal{M} \subset \mathcal{D}$
                 \lspstart
                 \STATE {\color{lspcolor}Sample perturbation $\mathbf{p}=(p_1,\ldots,p_q)\sim \mathcal{P}(k)$}
                 \STATE {\color{lspcolor}Set $\mathbf{z}\gets \mathbf{p}\circ \sqrt{\mathbf{e}}=(p_1\sqrt{e_1},\ldots,p_q\sqrt{e_q})$}
                 \lspend
        
                 \STATE For samples in $\mathcal{M}$, compute $\log\pi_\theta(a_t \mid s_t,\mathbf{z},\mathbf{V})$ via \autoref{alg:lsp_forward}
                 \STATE $r_t(\theta) \gets \exp\!\Big(\log\pi_\theta(a_t|s_t,\mathbf{z},\mathbf{V}) - \log\pi_{\theta_k}(a_t|s_t)\Big)$
                 \STATE $L^{\text{CLIP}}(\theta)\gets \frac{1}{|\mathcal{M}|}\sum \min\Big(r_t(\theta)\hat{A}_t,\ \text{clip}(r_t(\theta),1-\epsilon,1+\epsilon)\hat{A}_t\Big)$
                 \STATE Update actor: $\theta \leftarrow \arg\max_{\theta}\ L^{\text{CLIP}}(\theta)$
                 \STATE Update value: $\psi \leftarrow \arg\min_{\psi}\ \frac{1}{|\mathcal{M}|}\sum\big(V_\psi(s_t)-\hat{R}_t\big)^2$
              \ENDFOR
           \ENDFOR
    \end{algorithmic}
    \end{algorithm}
    \vspace{-1.5em}
        
    \begin{algorithm}[H]
        \caption{Policy forward with latent perturbation}
        \label{alg:lsp_forward}
        \small
        \begin{algorithmic}[1]\raggedright
        \INPUT encoder CNN, policy head MLP, image $o \in \mathbb{R}^{H \times W \times C}$, \lsp{perturbation scale $\mathbf{z} \in \mathbb{R}^{q}$, PCA directions $\mathbf{V} \in \mathbb{R}^{D \times q}$}
        \STATE Encode observation: $i \gets \text{CNN}(o) \in \mathbb{R}^{D}$
        \lspstart
        \STATE Sample base noise: $\varepsilon \sim \mathcal{N}(0, I_q)$
        \STATE Scale along PCA axes: $\Delta \gets \varepsilon \circ \mathbf{z} \in \mathbb{R}^{q}$
        \STATE Project to latent space: $\Delta^{\text{lat}} \gets \mathbf{V}\Delta \in \mathbb{R}^{D}$
        \STATE Perturb latent: $\hat{i} \gets i + \Delta^{\text{lat}}$
        \lspend
        \STATE Compute action distribution parameters: $(\mu,\Sigma) \gets \text{MLP}(\hat{i})$
        \STATE Evaluate log-likelihood of action $a$: $\log\pi_\theta(a\mid o,\mathbf{z},\mathbf{V}) \gets \log \mathcal{N}(a;\mu,\Sigma)$
        \OUTPUT $(\mu,\Sigma)$ and $\log\pi_\theta(a\mid o,\mathbf{z},\mathbf{V})$
        \end{algorithmic}
    \end{algorithm}
    \end{minipage}

\newpage
\section{Compute and Memory Overhead of LSP}
\label{app:lsp_overhead}

We quantify the additional systems cost introduced by Latent Space Perturbation (LSP) relative to the underlying RL baseline.
We report end-to-end training throughput and GPU memory usage, and then isolate the two LSP-specific contributors: (i) the actor forward pass under latent perturbation and (ii) the periodic PCA recomputation that defines the perturbation directions.

\subsection{End-to-end overhead (SAC vs.\ SAC+LSP)}
\label{app:lsp_overhead_end2end}
\begin{figure}[t!]
    \centering
    \includegraphics[width=0.7\linewidth]{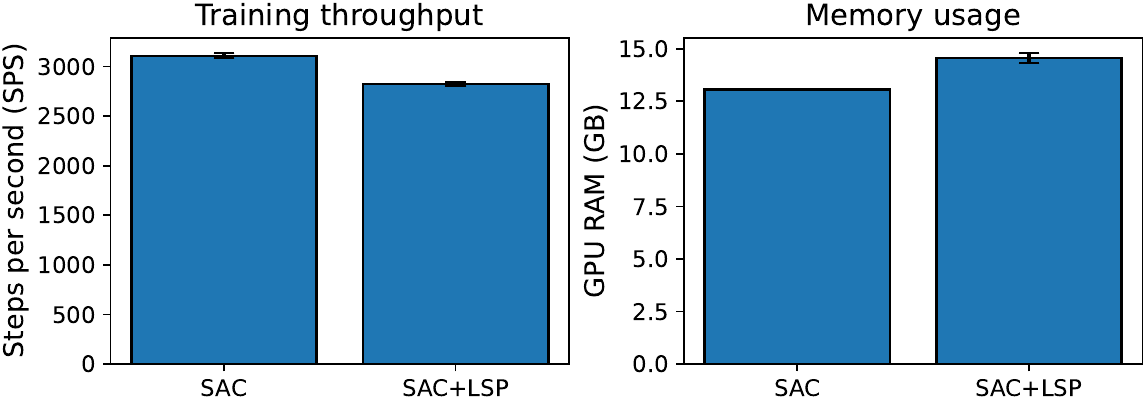}
    \vspace{-0.5em}
    \caption{
        \textbf{End-to-end training throughput and memory (SAC vs.\ SAC+LSP) on LSP-Bench.}
        LSP adds modest overhead: SPS drops by $\approx 9\%$ and mean GPU RAM rises by $\approx 11.5\%$, indicating it is lightweight relative to standard SAC updates.
    }
    \vspace{-2em}
    \label{fig:lsp_end_to_end_overhead}
\end{figure}

\autoref{tab:lsp_overhead_end2end} compares standard SAC to SAC augmented with LSP on LSP-Bench. We report steps per second (SPS) and average GPU RAM usage during training; \autoref{fig:lsp_end_to_end_overhead} visualizes the same comparison. 
Overall, LSP reduces end-to-end throughput by $\approx 9\%$ and increases average GPU RAM usage by $\approx 11.5\%$ on this benchmark.

\begin{table}[t]
    \centering
    \small
    \caption{End-to-end overhead of LSP on LSP-Bench (SAC).}
    \begin{tabular}{lccc}
    \toprule
    Metric & SAC & SAC+LSP & $\Delta$ \\
    \midrule
    SPS $\uparrow$ & $3107.67 \pm 26.42$ & $2824.33 \pm 22.37$ & $-9.11\%$ \\
    GPU RAM (GB) $\downarrow$ & $13.05 \pm 0.036$ & $14.55 \pm 0.235$ & $+11.49\%$ \\
    \bottomrule
    \end{tabular}
    \label{tab:lsp_overhead_end2end}
\end{table}

\subsection{Actor forward-pass overhead vs.\ number of perturbed directions $q$}
\label{app:lsp_overhead_actor}
\begin{figure}[t]
    \centering
    \includegraphics[width=0.48\linewidth]{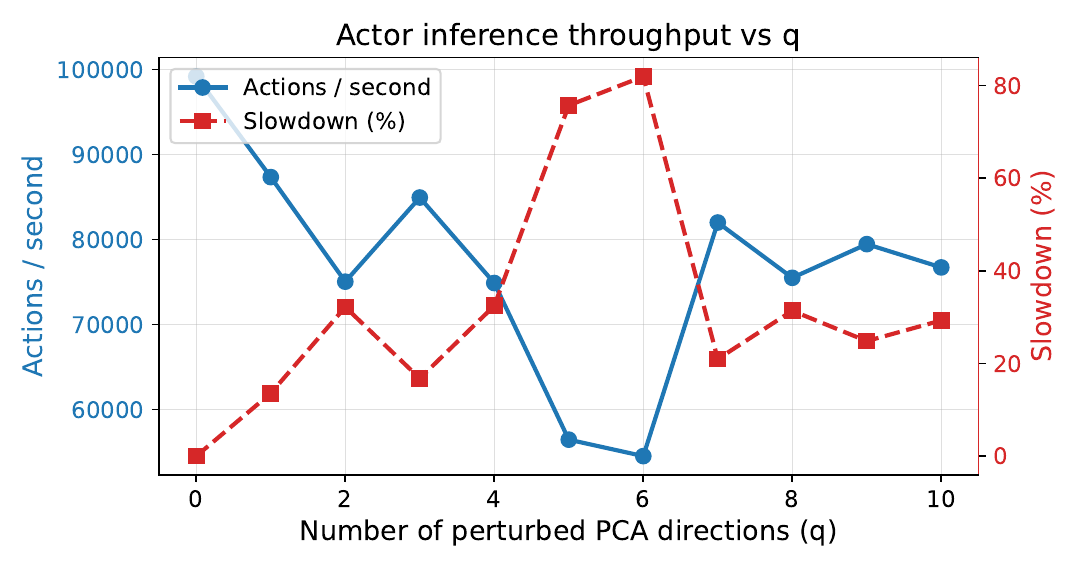}
    \caption{Actor throughput as a function of the number of perturbed directions $q$.}
    \label{fig:lsp_actor_throughput_vs_q}
\end{figure}

To isolate the incremental cost in the policy forward pass, we measure actor throughput (actions/s) as a function of the number of perturbed PCA directions $q$.
\autoref{tab:lsp_overhead_actor} reports throughput and slowdown relative to the unperturbed actor ($q=0$), with \autoref{fig:lsp_actor_throughput_vs_q} providing the corresponding plot.

\begin{table}[t]
\centering
\small
\caption{Actor throughput under latent perturbation vs.\ the number of perturbed PCA directions $q$ (slowdown computed w.r.t.\ $q=0$).}
\begin{tabular}{rcc}
\toprule
$q$ & actions/s $\uparrow$ & slowdown $\downarrow$ \\
\midrule
0  & 99,190 & 0.00\% \\
1  & 87,345 & 13.58\% \\
2  & 75,037 & 32.19\% \\
3  & 84,941 & 16.78\% \\
4  & 74,880 & 32.47\% \\
5  & 56,443 & 75.73\% \\
6  & 54,507 & 81.98\% \\
7  & 82,001 & 20.96\% \\
8  & 75,492 & 31.39\% \\
9  & 79,455 & 24.84\% \\
10 & 76,715 & 29.30\% \\
\bottomrule
\end{tabular}

\label{tab:lsp_overhead_actor}
\end{table}

The slowdown increases with $q$ as expected, since larger $q$ adds latent-space sampling and a low-rank projection.
The small non-monotonicity across adjacent $q$ values is typical for GPU workloads at these matrix sizes and does not change the end-to-end conclusion in \autoref{tab:lsp_overhead_end2end}.

\subsection{Cost of PCA recomputation}
\label{app:lsp_overhead_pca}
\begin{figure}[t]
    \centering
    \includegraphics[width=0.7\linewidth]{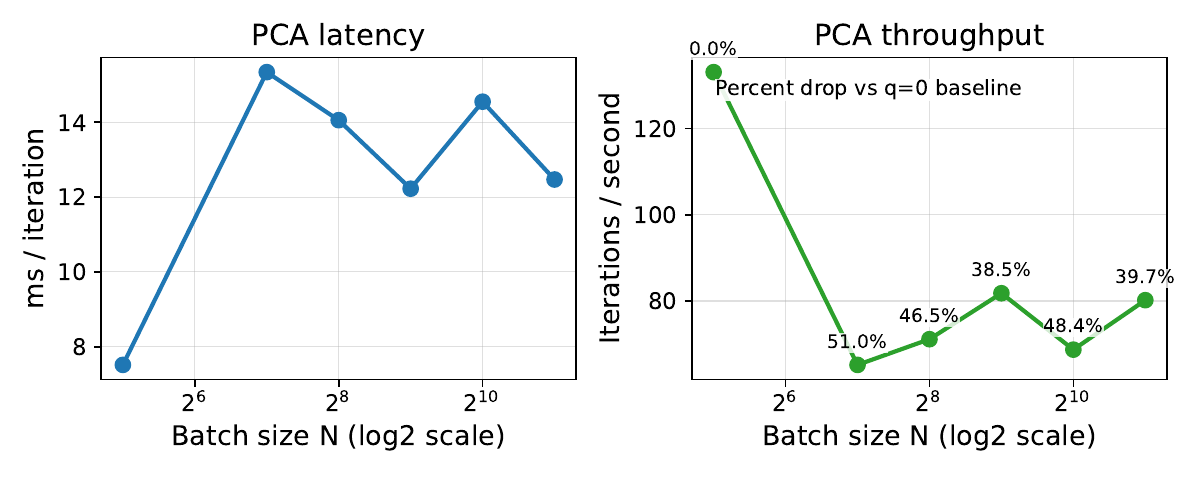}
    \caption{PCA recomputation cost for latent matrices of shape $(N,256)$ (top-$q=10$ components).}
    \label{fig:lsp_pca_cost}
\end{figure}

LSP periodically recomputes a PCA from encoder latents.
We benchmark PCA on latent matrices of shape $(N,256)$ while extracting the top $q=10$ components.
\autoref{tab:lsp_overhead_pca} reports runtime and throughput, and \autoref{fig:lsp_pca_cost} visualizes scaling with $N$.

\begin{table}[t]
\centering
\small
\caption{PCA benchmark on latent batches of size $N$ (latent dim.\ 256, extracting $q=10$ components).}
\begin{tabular}{rccc}
\toprule
$N$ & ms / PCA $\downarrow$ & PCA iters/s $\uparrow$ & drop \\
\midrule
32   & 7.5145  & 133.08 & 0.00\% \\
128  & 15.3417 & 65.18  & 51.03\% \\
256  & 14.0577 & 71.14  & 46.54\% \\
512  & 12.2255 & 81.80  & 38.53\% \\
1024 & 14.5511 & 68.72  & 48.36\% \\
2048 & 12.4709 & 80.19  & 39.74\% \\
\bottomrule
\end{tabular}

\label{tab:lsp_overhead_pca}
\end{table}

At the latent dimensionalities used in this work, PCA is inexpensive in absolute terms and is amortized by performing recomputation only periodically. 
In deployments with tighter compute budgets, overhead can be reduced by increasing the PCA period $f$ (less frequent recomputation) and/or using smaller $q$ (cheaper actor-side perturbation; Table~\ref{tab:lsp_overhead_actor} and \autoref{fig:lsp_actor_throughput_vs_q}).

\subsection{Discussion}
\label{app:lsp_overhead_discussion}
On LSP-Bench with SAC, LSP incurs a $\approx 9\%$ reduction in end-to-end throughput and a $\approx 11.5\%$ increase in average GPU RAM usage (\autoref{tab:lsp_overhead_end2end}, \autoref{fig:lsp_end_to_end_overhead}).
In the main paper, this overhead corresponds to substantially improved robustness under appearance-only shifts. When OOD visual generalization is a target requirement, this trade-off is favorable; when the visual distribution is stationary, the baseline may be preferred for slightly higher throughput.

\subsection{Generalization Gap}
\label{app:procgen_generalization_gap}

\begin{figure}[H]
    \centering
    \includegraphics[width=1\linewidth]{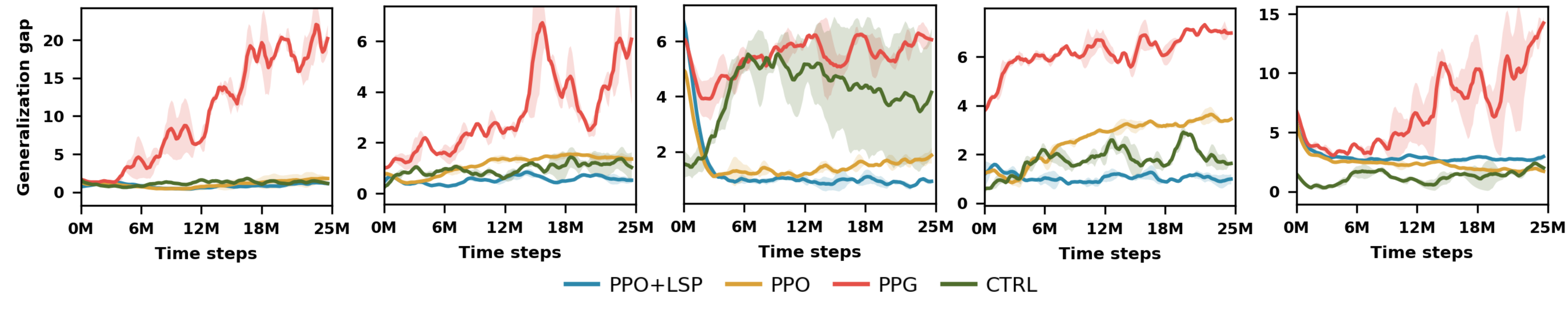}
    \caption{Generalization Gap of PPO+LSP, PPO, PPG and CTRL baselines on the Procgen benchmark. }
    \label{fig:gen_gap}
\end{figure}

\section{Higher-Order Structure of the Train-to-Evaluation Latent Shift}
\label{app:higher_order_shift}

Our analysis does not assume Gaussian latent marginals. Instead, LSP relies on the weaker observation that train-to-evaluation shifts along dominant training directions are largely characterized by changes in location and scale. We quantify this using the Wasserstein-1 distance, normalized by the training standard deviation, before normalization, after removing the mean shift, and after additionally matching the variance.

Aggregated over the top-10 principal components and weighted by their explained training variance, the total discrepancy is approximately $0.60$. Mean shifts account for about $67\%$, variance shifts for an additional $7\%$, and approximately $25\%$ remains as a higher-order residual. The split-half distance between independent subsets of the training distribution is only approximately $0.007$, confirming that this residual is well above the sampling floor.

\begin{table}[t]
\centering
\caption{Variance-weighted decomposition of the train-to-evaluation distribution shift over the top-10 training principal components. The residual denotes the discrepancy remaining after separately removing mean and variance differences.}
\label{tab:wasserstein_decomposition}
\begin{tabular}{lc}
\toprule
Shift component & Fraction of total discrepancy \\
\midrule
Mean shift & $67\%$ \\
Variance shift & $7\%$ \\
Higher-order residual & $25\%$ \\
\bottomrule
\end{tabular}
\end{table}

The decomposition is strongly component-dependent. PC0 is dominated by a mean shift, whereas PC5 mainly changes in scale, with evaluation variance increasing by approximately $1.4\times$. PC9 exhibits comparable contributions from both. Thus, no single moment explains all dominant directions.

The remaining residual does not indicate the emergence of new dominant modes. The marginals remain single-peaked, while the residual mainly reflects higher-order changes in shape. For example, PC0 has excess kurtosis of approximately $-1.2$, while PC1 has skewness of approximately $-0.9$.

These results clarify the assumption behind LSP. The method neither requires Gaussian latent distributions nor attempts to reproduce the full evaluation-time distribution. It assumes only that OOD variation remains concentrated along dominant training directions and that a substantial part of the shift is captured by changes in location and scale. LSP therefore expands support along these directions using variance-scaled perturbations.

The higher-order residual also marks a limitation: LSP does not explicitly model changes in skewness, kurtosis, or other higher-order statistics. Nevertheless, because most of the observed shift remains aligned with the training subspace and is explained by low-order changes, perturbing these directions can improve robustness without explicitly modeling the complete OOD distribution.

\section{Cross-Architecture Validation of Latent Geometry under Naturalistic Visual Shifts}
\label{app:cross_architecture_geometry}

To test whether the observed latent-space geometry is specific to CNN representations, we repeat the analysis with a frozen DINOv2 Vision Transformer. We evaluate whether OOD visual changes remain concentrated within the dominant training directions on both RL-ViGen and the Distracting Control Suite (DCS), where evaluation uses naturalistic real-world image and video backgrounds.

\paragraph{RL-ViGen.}
We analyze RL-ViGen Robosuite Door observations under the training distribution and evaluation settings with video-background shifts. Despite the different architecture and pretraining procedure, the DINOv2 representation remains strongly low-rank: the first ten training principal components explain approximately $82\%$ of the total latent variance.

The train and evaluation principal subspaces remain closely aligned, with

$$
\cos\theta_{0} \approx 0.995,
\qquad
\cos\theta_{30} \approx 0.95.
$$

The distribution shift is also concentrated in the leading training directions. The fraction of variance-shift energy captured by these components is approximately

$$
E_{\lambda}^{K} \approx 0.99,
$$

while the mean shift is similarly concentrated within the dominant training subspace. These properties remain qualitatively stable as shift severity increases, indicating that the visual shift primarily changes statistics along existing latent directions rather than introducing new dominant ones.

\paragraph{Distracting Control Suite.}
We further analyze Cartpole Balance and Cheetah Walk under Easy, Medium, and Hard DCS shifts using both the task-trained CNN and frozen DINOv2 ViT.

For the CNN, the first ten training principal components explain approximately $90\%$ of the variance on Cartpole Balance and $94\%$ on the Cheetah task. Across all evaluation severities, the shift remains concentrated within these directions. The top-ten mean-shift energy lies approximately in the range

$$
E_{\mu}^{10} \approx 0.94\text{--}0.98,
$$

while the corresponding variance-shift concentration remains close to

$$
E_{\lambda}^{10} \approx 1.00.
$$

The principal subspaces also remain highly aligned:

$$
\cos\theta_{0} \geq 0.999,
$$

with deeper alignment remaining high,

$$
\cos\theta_{30} \approx 0.86\text{--}0.98.
$$

Thus, even under real-world video backgrounds, the task-trained CNN primarily expresses the new variability along directions already dominant during training.

The DINOv2 representation shows the same qualitative behavior, although it weakens with shift severity. Its first ten training components explain approximately $79\%$ of the variance on Cartpole Balance and $77\%$ on the Cheetah task. Under the Easy setting, the shift remains strongly concentrated in the dominant training directions, with

$$
E_{\lambda}^{10} \approx 0.94\text{--}0.98,
\qquad
\cos\theta_{30} \approx 0.98.
$$

The leading directions are also almost perfectly aligned, with $\cos\theta_{0}$ approaching $0.999$.

Under stronger video shifts, however, the DINOv2 representation increasingly expresses OOD variation outside the first ten training components. This effect is weaker for the task-trained CNN, suggesting that task-specific representation learning helps preserve the dominant latent subspace under severe visual changes.

\paragraph{Implications for the LSP assumption.}
These experiments show that the latent geometry motivating LSP is not specific to CNNs. Both the task-trained CNN and frozen DINOv2 ViT exhibit low-dimensional training representations, and under mild-to-moderate shifts, OOD variability remains concentrated within dominant training directions.

At the same time, DCS reveals an important limitation: sufficiently strong naturalistic shifts can cause a frozen general-purpose encoder to occupy directions outside the dominant training subspace. We therefore do not assume that arbitrary OOD shifts remain confined to a fixed low-dimensional subspace. Rather, LSP targets the regime where most OOD variation remains aligned with dominant training directions, which is observed most consistently for task-trained representations.

\section{Robustness to Perturbation Strategy and Interpretation of $p_{\max}$}
\label{app:lsp_hyperparameter_sensitivity}

LSP introduces two related design choices: the perturbation scheduling strategy and the maximum perturbation magnitude $p_{\max}$. We perform additional experiments to separate their effects and clarify what determines an appropriate perturbation strength.

\paragraph{Robustness to the perturbation scheduling strategy.}
To isolate the effect of the scheduling strategy, we fix the perturbation magnitude to $1.0$ and evaluate all six strategies considered in the paper.

Table~\ref{tab:fixed_magnitude_strategies} shows that the strategies generally form a relatively tight performance cluster. Strategy~5 performs best at several intermediate checkpoints, while other strategies become competitive at different stages; for example, Strategy~1 achieves the highest mean return at 9M steps. Thus, no single scheduling strategy consistently dominates throughout training.

\begin{table*}[t]
\centering
\caption{
Ablation over the six LSP perturbation scheduling strategies with the perturbation magnitude fixed to $1.0$. Values report mean episodic return $\pm$ standard deviation. Fixing the magnitude isolates the effect of the temporal scheduling profile from the effect of perturbation strength.
}
\label{tab:fixed_magnitude_strategies}
\resizebox{\textwidth}{!}{
\begin{tabular}{lcccccccccc}
\toprule
Strategy &
0M & 1M & 2M & 3M & 4M & 5M & 6M & 7M & 8M & 9M \\
\midrule
Strategy 1 &
$-86.1\pm2.6$ &
$-37.4\pm23.6$ &
$1.7\pm6.8$ &
$37.5\pm3.5$ &
$43.9\pm10.5$ &
$52.0\pm13.3$ &
$46.0\pm16.4$ &
$47.4\pm10.7$ &
$47.8\pm6.3$ &
$\mathbf{55.8\pm9.4}$ \\

    Strategy 2 &
    $-76.1\pm0.8$ &
    $15.0\pm3.4$ &
    $\mathbf{26.0\pm4.3}$ &
    $34.4\pm3.6$ &
    $36.4\pm6.2$ &
    $29.2\pm3.2$ &
    $39.5\pm4.9$ &
    $32.3\pm0.9$ &
    $25.5\pm12.4$ &
    $25.0\pm10.8$ \\

    Strategy 3 &
    $-87.2\pm0.5$ &
    $\mathbf{16.1\pm1.6}$ &
    $22.1\pm8.9$ &
    $38.0\pm7.4$ &
    $35.2\pm12.7$ &
    $46.3\pm8.9$ &
    $41.6\pm9.2$ &
    $26.6\pm17.9$ &
    $39.2\pm6.6$ &
    $47.4\pm6.5$ \\

    Strategy 4 &
    $-85.4\pm15.4$ &
    $11.8\pm1.1$ &
    $12.9\pm13.1$ &
    $38.1\pm6.8$ &
    $37.3\pm16.4$ &
    $43.5\pm20.7$ &
    $46.8\pm17.8$ &
    $44.9\pm16.9$ &
    $6.6\pm42.1$ &
    $19.5\pm31.9$ \\

    Strategy 5 &
    $-80.8\pm3.7$ &
    $8.7\pm3.4$ &
    $25.1\pm7.0$ &
    $\mathbf{41.4\pm5.1}$ &
    $\mathbf{51.7\pm1.7}$ &
    $\mathbf{53.3\pm4.3}$ &
    $\mathbf{55.4\pm0.1}$ &
    $\mathbf{49.4\pm2.7}$ &
    $\mathbf{50.3\pm0.8}$ &
    $43.7\pm4.1$ \\

    Strategy 6 &
    $-98.6\pm6.6$ &
    $14.9\pm3.2$ &
    $13.1\pm13.3$ &
    $37.5\pm6.4$ &
    $33.9\pm14.1$ &
    $40.8\pm18.9$ &
    $44.4\pm16.2$ &
    $37.5\pm12.0$ &
    $1.5\pm38.7$ &
    $14.1\pm28.3$ \\
    \bottomrule
\end{tabular}
}

\end{table*}

These results suggest that the perturbation magnitude matters more than the exact temporal schedule. Once magnitude is controlled, the strategies do not exhibit a single sharply isolated optimum, indicating limited sensitivity to the scheduler choice.

\paragraph{What determines the appropriate $p_{\max}$?}
A natural hypothesis is that $p_{\max}$ should be smaller in environments with more visual distractors. Our results instead suggest that it is more closely related to how sensitive the policy is to perturbations along dominant latent directions.

For each Procgen environment, we perturb the leading training principal component with amplitude $0.9$ training standard deviations and measure the KL divergence between the original and perturbed policies, denoted $\mathrm{KL}@0.9$. Larger values indicate that the perturbed direction carries more task-relevant information.

The resulting measurements are reported in Table~\ref{tab:pmax_policy_sensitivity}.

\begin{table}[t]
\centering
\caption{
Relationship between the tuned perturbation magnitude $p_{\max}$, policy sensitivity to perturbation of the leading training principal component, and the number of visual distractors in the five Procgen environments. Policy sensitivity is measured as the KL divergence between the original and perturbed policy outputs for a perturbation magnitude of $0.9$ training standard deviations.
}
\label{tab:pmax_policy_sensitivity}
\begin{tabular}{lccc}
\toprule
Environment & $p_{\max}$ & $\mathrm{KL}@0.9$ & Distractors \\
\midrule
Bigfish   & $0.80$ & $0.014$ & $4$ \\
Dodgeball & $1.10$ & $0.005$ & $5$ \\
Heist     & $0.80$ & $0.062$ & $1$ \\
Jumper    & $1.10$ & $0.017$ & $1$ \\
Plunder   & $1.10$ & $0.006$ & $3$ \\
\bottomrule
\end{tabular}
\end{table}

Across the five environments, the tuned $p_{\max}$ is negatively correlated with policy sensitivity:

$$
r=-0.68.
$$

By contrast, its correlation with the number of visual distractors is weak:

$$
r\approx +0.15,
$$

suggesting that distractor count alone does not determine the preferred perturbation strength.

Heist illustrates this most clearly. Its policy sensitivity is substantially larger than in the other environments,

$$
\mathrm{KL}@0.9 = 0.062,
$$

compared with $0.005$--$0.017$ elsewhere, and it uses the lower value $p_{\max}=0.8$. This suggests that $p_{\max}$ is governed primarily by how much task-relevant information is carried by the perturbed latent directions.

\paragraph{Interpretation.}
A dominant principal component may encode both nuisance variation and task-relevant information. Stronger perturbations therefore improve robustness only until they begin to disrupt information used by the policy. The policy KL response provides a direct measure of this trade-off: directions that produce larger policy changes should be perturbed more conservatively.

Because this analysis contains only five Procgen environments, the correlation should be interpreted as a mechanistic trend rather than a statistically established relationship. Nevertheless, it provides a practical interpretation of $p_{\max}$ and suggests policy sensitivity to PCA-aligned perturbations as a useful diagnostic for selecting perturbation strength.

\section{Scalability to High-Dimensional Latent Spaces}
\label{app:pca_scalability}

LSP periodically estimates the dominant principal directions of the training latent distribution. To test whether this becomes expensive for high-dimensional representations, we benchmark PCA recomputation as the latent dimensionality $D$ increases.

LSP requires only the leading $q$ principal directions rather than a full eigendecomposition. We therefore use truncated randomized PCA, whose complexity for fixed sample count $N$ and number of retained components $q$ is approximately

$$
\mathcal{O}(NDq),
$$

and thus scales linearly with $D$.

We benchmark PCA recomputation from $D=256$ to $D=8192$, keeping $N$ and $q$ fixed to their training values. Results are shown in Table~\ref{tab:pca_scalability}.

\begin{table}[t]
\centering
\caption{
PCA recomputation time as a function of the latent dimensionality $D$. The number of samples $N$ and retained principal components $q$ are held fixed. The final column reports the runtime relative to $D=256$.
}
\label{tab:pca_scalability}
\begin{tabular}{ccc}
\toprule
Latent dimension $D$ & PCA time (ms) & Relative time \\
\midrule
$256$  & $11.73$ & $1.00\times$ \\
$512$  & $12.09$ & $1.03\times$ \\
$1024$ & $10.41$ & $0.89\times$ \\
$2048$ & $12.95$ & $1.10\times$ \\
$4096$ & $12.57$ & $1.07\times$ \\
$8192$ & $14.71$ & $1.25\times$ \\
\bottomrule
\end{tabular}
\end{table}

Increasing $D$ by a factor of $32$, from $256$ to $8192$, increases recomputation time only from $11.73$\,ms to $14.71$\,ms, corresponding to approximately $1.25\times$. Across this range, PCA recomputation therefore remains on the order of $10$--$15$\,ms on our hardware.

The PCA basis is recomputed only once every $f$ training steps and reused between updates, further amortizing this cost. The $D=1024$ regime is already used in our RL-ViGen experiments, demonstrating that LSP is applicable to substantially larger latent representations than those used in LSP-Env and Procgen. Separately, our end-to-end measurements on LSP-Bench show approximately a $9\%$ throughput reduction and an $11.5\%$ increase in GPU memory usage relative to the baseline (see~\autoref{app:lsp_overhead}).

Overall, PCA recomputation does not become a practical bottleneck over the tested dimensionalities: increasing $D$ from $256$ to $8192$ produces only a small increase in absolute recomputation time, which is further amortized by periodic updates.

Finally, LSP applies PCA to the single latent representation consumed by the policy head. Applying perturbations to per-token Key or Value representations inside a Transformer would constitute a different method and is outside the scope of this work.

\section{Does LSP Require Disentangled Representations?}
\label{app:disentanglement}

LSP does not assume that task-relevant information and nuisance visual factors are disentangled into separate latent directions. It neither \emph{induces} a disentangled representation nor \emph{requires} the encoder to have already separated these factors.

\paragraph{LSP does not explicitly induce disentanglement.}
LSP operates between the visual representation and the policy head, perturbing dominant directions of the training latent distribution during policy optimization. It introduces no disentanglement objective, assigns no semantic meaning to individual principal components, and does not require PCs to correspond to independent generative factors.

Its goal is therefore not to learn a factorized representation, but to make the policy robust to structured variation along dominant training directions.

\paragraph{Individual principal components need not be purely task-related or purely nuisance-related.}
LSP only requires that a substantial fraction of the OOD-induced shift lies within the dominant training subspace. In our experiments, the concentration metric $E^K$ typically lies in the range $0.85$--$0.99$. This does not imply that individual PCs contain exclusively task or nuisance information.

A leading PC may contain both. Perturbing such a direction can therefore also disturb task-relevant signal, and $p_{\max}$ controls this trade-off. This is consistent with the policy-sensitivity analysis in Appendix~\ref{app:lsp_hyperparameter_sensitivity}: directions that produce larger policy changes require smaller perturbation magnitudes.

\paragraph{Mixed task and nuisance information in LSP-Env.}
LSP-Env Protocol~3 provides a mixed-feature example. Here, the nuisance factor is the agent's appearance, which is inseparable in pixel space from task-relevant information such as its position. Thus, nuisance information need not occupy separate ``style-only'' PCs.

Nevertheless, LSP improves OOD performance in this protocol. Its preferred perturbation magnitude is lower than in Protocol~1, with $p_{\max}=0.8$ versus $p_{\max}=0.9$, consistent with more conservative perturbation when dominant directions contain more task-relevant information.

\paragraph{Required assumption.}
The actual assumption of LSP is therefore weaker than disentanglement: OOD-induced variation should substantially overlap with the dominant training-time latent subspace.

PCA identifies directions of large statistical variation rather than semantic factors. LSP therefore does not require individual PCs to correspond to background, appearance, or task state. It exploits only the geometric observation that visual OOD shifts often remain concentrated along directions already present during training.

When this condition holds, perturbing those directions exposes the policy to structured variation resembling the OOD shift. When substantial energy appears in previously unseen directions, the assumption weakens and the benefit of LSP is expected to decrease, consistent with the severe-shift analyses elsewhere in the paper.

\end{document}

%% file: math_commands.tex
\usepackage{amsmath,amsfonts,bm}

\def\eqref#1{equation~\ref{#1}}

\def\1{\bm{1}}

\DeclareMathAlphabet{\mathsfit}{\encodingdefault}{\sfdefault}{m}{sl}
\SetMathAlphabet{\mathsfit}{bold}{\encodingdefault}{\sfdefault}{bx}{n}

